\pdfoutput=1

\documentclass[11pt]{article}

\usepackage[preprint]{acl}

\usepackage{times}
\usepackage{latexsym}
\usepackage{multirow}
\usepackage{array}

\usepackage{amsmath}
\usepackage{amssymb}
\usepackage{mathtools}
\usepackage{amsthm}
\usepackage{caption}
\usepackage{booktabs} 
\usepackage{algorithm}
\usepackage{algorithmic}
\usepackage[T1]{fontenc}

\usepackage[utf8]{inputenc}

\usepackage{microtype}

\usepackage{inconsolata}

\usepackage{graphicx}

\definecolor{Green}{rgb}{0.0, 0.5, 0.0}
\definecolor{Red}{rgb}{0.6, 0.0, 0.0}

\title{\texorpdfstring{{NodeRAG: \\Structuring Graph-based RAG with Heterogeneous Nodes}}}

\author{Tianyang Xu$^{1}$, Haojie Zheng$^{2}$, Chengze Li$^{1}$, Haoxiang Chen$^{1}$\\ \textbf{Yixin Liu$^{3}$, Ruoxi Chen, Lichao Sun$^{3}$} \\
  $^{1}$ Columbia University, $^{2}$ University of Pennsylvania, $^{3}$ Lehigh University \\
  \texttt{tx2240@columbia.edu}, \quad \texttt{haojiez@seas.upenn.edu} 
}

\begin{document}
\maketitle
\begin{abstract}
Retrieval-augmented generation (RAG) empowers large language models to access external and private corpus, enabling factually consistent responses in specific domains. By exploiting the inherent structure of the corpus, graph-based RAG methods further enrich this process by building a knowledge graph index and leveraging the structural nature of graphs. However, current graph-based RAG approaches seldom prioritize the design of graph structures. Inadequately designed graph not only impede the seamless integration of diverse graph algorithms but also result in workflow inconsistencies and degraded performance. To further unleash the potential of graph for RAG, we propose {NodeRAG}, a graph-centric framework introducing heterogeneous graph structures that enable the seamless and holistic integration of graph-based methodologies into the RAG workflow. By aligning closely with the capabilities of LLMs, this framework ensures a fully cohesive and efficient end-to-end process. Through extensive experiments, we demonstrate that NodeRAG exhibits performance advantages over previous methods, including GraphRAG and LightRAG, not only in indexing time, query time, and storage efficiency but also in delivering superior question-answering performance on multi-hop benchmarks and open-ended head-to-head evaluations with minimal retrieval tokens. Our GitHub repository could be seen at this \href{https://github.com/Terry-Xu-666/NodeRAG}{link}.
\end{abstract}

\setlength{\abovecaptionskip}{-2pt}  
\setlength{\belowcaptionskip}{-5pt} 
\begin{figure*}[t]
    \centering
    \includegraphics[width=\textwidth]{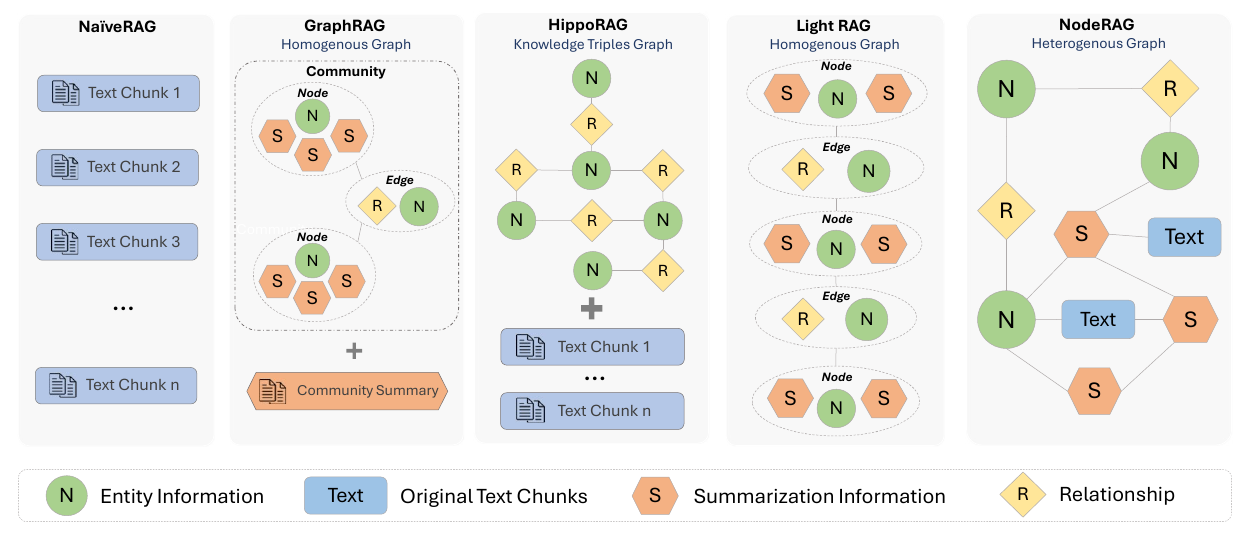} 
    \setlength{\abovecaptionskip}{0.05cm}
    \captionsetup{font={small}}
    \caption{Comparsions between NodeRAG and other RAG systems. {NaïveRAG} retrieving fragmented text chunks, leads to redundant information. {HippoRAG} introduces knowledge graphs but lacks high-level summarization. {GraphRAG} retrieves community summaries but may still produce coarse-grained information. {LightRAG} incorporates one-hop neighbors but retrieves redundant nodes. In contrast, {NodeRAG} utilizes multiple node types, including high-level elements, semantic units, and relationships, enabling more precise, hierarchical retrieval while reducing irrelevant information.}
    \label{fig:comparison}
\end{figure*}

\section{Introduction}

Retrieval-augmented generation (RAG) has emerged as a solution to the challenges posed by the rapid evolution of real-world knowledge domains \cite{fan2024survey}, coupling large language models (LLMs) with an external retrieval mechanism to ensure the generation of factually consistent and contextually relevant information \cite{tonmoy2024comprehensive,shrestha2024fairrag,liu2024lost}. Despite recent progress, current RAG methods face notable shortcomings in handling multi-hop reasoning \cite{luo2023reasoning,wang2024knowledge} and summary-level queries \cite{han2024retrieval,wen2023mindmap} due to their insufficient utilization of data structures and lack of high-level understanding of the text corpus. Graph-based RAG methods \cite{tian2024graph,park2023graph} have been proposed to enhance retrieval and question-answering performance, specifically addressing the two main challenges faced by traditional RAG approaches. Leveraging LLMs to decompose raw data into graph structures \cite{gutierrez2024hipporag,he2024g} for utilizing structural information, as well as employing LLMs for summary-based enhancements \cite{edge2024local,guo2024lightrag} to derive insights beyond the original text, has gradually become mainstream approaches.

However, previous Graph-based RAG works \cite{trajanoska2023enhancing,gutierrez2024hipporag} have rarely considered the critical role of graph structures, i.e., what forms of graph better support RAG. Among existing approaches, knowledge graphs \cite{sanmartin2024kg,wang2024knowledge} extract triples, with the graph containing only structural information, yet retrieval context remains confined to text chunks, which often lack semantic coherence and include unrelated information. While current methods attempt to incorporate more information into the graph and extract deeper insights, they suffer from inefficiencies and inconsistencies due to inadequately designed structures. For instance, as illustrated in Figure \ref{fig:comparison}, GraphRAG \cite{edge2024local}, adopt a tightly coupled entity-event homogeneous structure, hindering the integration of original context and summary information into the graph. This results in inconsistencies in retrieval methods (separating local and global retrieval) and leads to coarse-grained retrieval, where retrieving an entity indiscriminately includes all associated events, adding irrelevant information.

To address these limitations, we propose NodeRAG, which is built around a well-designed Heterogeneous Graph, comprehensively considering the entire process of graph indexing and searching, enabling fine-grained retrieval. The heterograph adheres to the principle of unfolding and flattening, decomposing different types of information to construct a heterogeneous fully nodalized graph where nodes serve distinct functions and roles. This means that entities, relationships, original text chunks, independently decomposed events from text chunks, and summaries extracted by LLMs are all represented as nodes within the graph. The heterograph not only encapsulates information from the original corpus but also extends beyond it, incorporating enriched insights such as key node attributes, and high-level discoveries. Each node in heterograph consists of unstructured content, while preserving structural connections between nodes, striking a balance between structural integrity and flexibility. As illustrated in Figure \ref{fig:comparison}, for a multi-hop question, NodeRAG can retrieve a semantically coherent, independent event (semantic unit) and high-level discoveries (high-level elements) related to key entities such as Harry, Neville, and the three-headed dog using graph algorithms, providing explainable and fine-grained retrievals as well as high-level understanding.

The key contributions of our work can be summarized in three main aspects. 

\noindent\textbf{(1) Better Graph Structure for RAG}~~The graph structure serves as the foundation for graph-based RAG where significance has been overlooked. Our work emphasizes its importance and introduces a graph structure that better supports RAG. 

\noindent\textbf{(2) Fine-grained and Explainable Retrieval}~~The heterograph enables fine-grained and functionally distinct nodes, allowing graph algorithms to effectively and reasonably identify key multi-hop nodes. This leads to more relevant retrieval with minimal retrieval context, enhancing both precision and interoperability. 

\noindent\textbf{(3) Unified-Level Information Retrieval}~~Decomposed information from documents and extracted insights from LLMs are not treated as separate layer but are instead unified as nodes within the heterograph. This integration allows for a cohesive framework capable of handling information needs across different levels. 

In addition, extensive experiments demonstrate that NodeRAG not only outperforms previous graph-based RAG methods on multi-hop tasks but also exhibits superior performance in open-ended head-to-head evaluations. With minimal retrieval tokens, it achieves highly precise retrieval while also demonstrating system-level efficiency advantages, including improvements in indexing time, query time, and storage efficiency, as shown in appendix \ref{RAGComparison}.

\section{NodeRAG}

The NodeRAG pipeline is built on a foundational graph structure defined as the heterograph, which will be introduced in Section \ref{NodeGraph}. The workflow is divided into two primary stages, graph indexing and graph searching. Graph indexing comprises three components, graph decomposition, graph augmentation, and graph enrichment, which are discussed in Sections \ref{Decomposition}, \ref{augmentation}, and \ref{enrichment}, respectively. This stage integrates various types of nodes and edges into the heterograph by leveraging LLMs and graph algorithms. The subsequent stage, graph searching, is detailed in Section \ref{searching} and combines the structural advantages of the heterograph with graph algorithms to efficiently retrieve relevant information. Moreover, the fundamental concepts and implementation details of the graph algorithms used in the pipeline are provided in Appendix \ref{Algorithms}, while the prompting instructions for LLMs can be found in Appendix \ref{Prompting in NodeRAG} for reference.

\begin{figure*}[t]
    \centering
    \includegraphics[width=\textwidth]{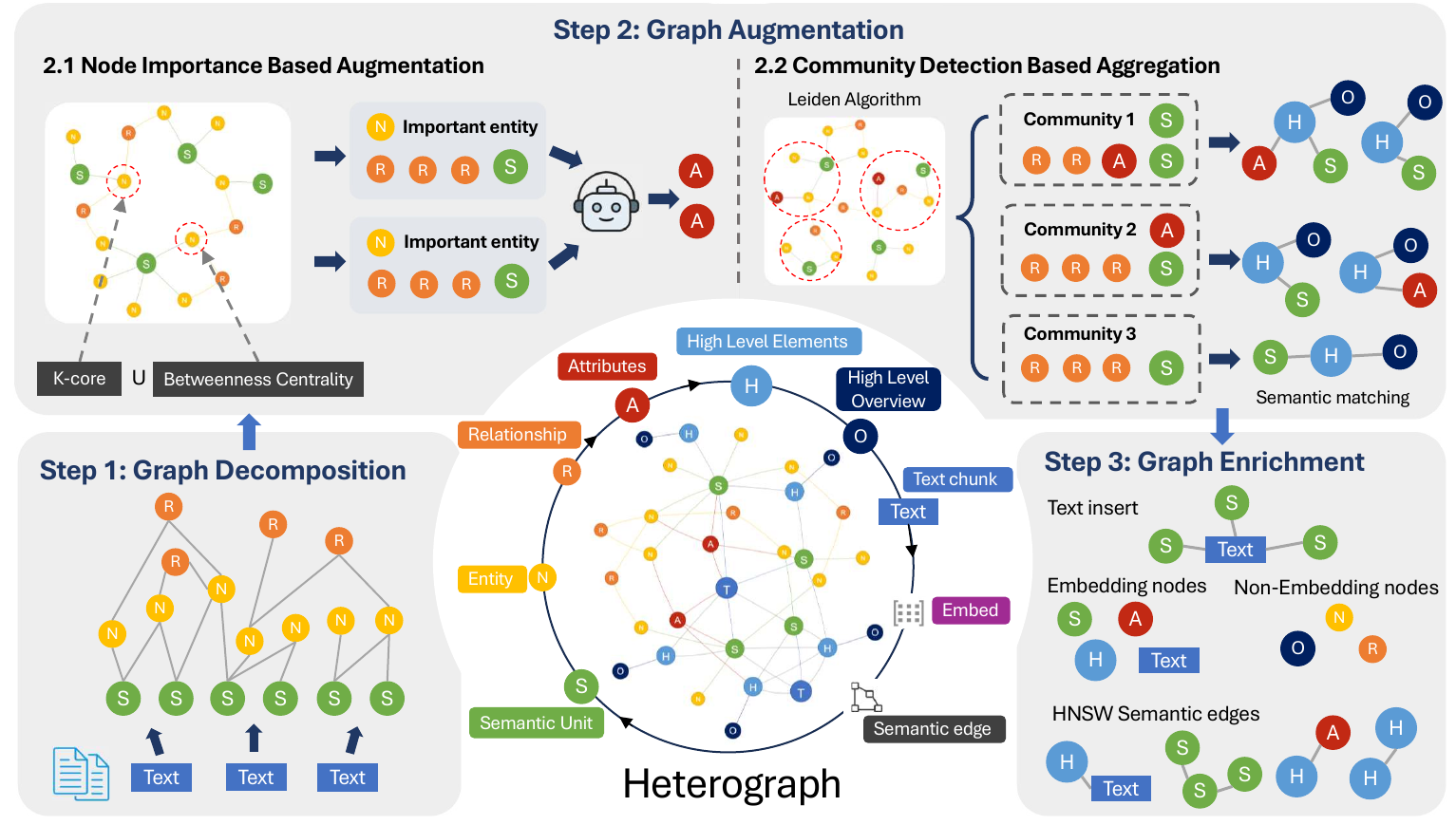} 
    \setlength{\abovecaptionskip}{0.1cm}
    \caption{Main indexing workflow of NodeRAG. It illustrates the step-by-step construction of the heterograph, including the process of graph decomposition, graph augmentation, and graph enrichment}
    \label{fig:process1}
\end{figure*}

\subsection{Heterograph}
\label{NodeGraph}
The concept of the heterograph embodies the principle of comprehensive unfolding and flattening of information into a fully nodalized structure. This structure achieves its granularity through the integration of seven hetero node types: entity (\(N\)), relationship (\(R\)), semantic unit (\(S\)), attribute (\(A\)), high-level elements (\(H\)), high-level overview (\(O\)), and text (\(T\)). Each node type is tailored to represent specific roles and characteristics of the information, enabling a fine-grained and functional decomposition of data. Mathematically, the heterograph is defined as:
\setlength{\abovedisplayskip}{2pt} 
\setlength{\belowdisplayskip}{2pt}
\[
\mathcal{G} = (\mathcal{V}, \mathcal{E}, \Psi),
\]
where \(\mathcal{G}\) is the heterograph, \( \mathcal{V} \) represents the set of nodes, \( \mathcal{E} \) is the set of edges, and \( \Psi : \mathcal{V} \to \mathrm{Types} \) is a mapping function that assigns each node \( v \in \mathcal{V} \) to a specific type. The set of node types, corresponds to the seven predefined types:
\[
\mathrm{Types} = \{N, R, S, A, H, O, T\}.
\] 
For any node \(v\), \( \Psi(v) \) defines its type, with each node type performing a distinct and well-defined function, as detailed in subsequent sections and appendix \ref{Algorithms}. For each \(e \in \mathcal{E}\), the default weight of \(e\) is set to 1, representing a basic connection between two nodes. Furthermore, we define \(\mathcal{V}_{\mathrm{types}}\) as the subset of nodes corresponding to a subset set \(\mathrm{types} \subseteq \mathrm{Types}\), formally expressed as:  
\[
\mathcal{V}_{\mathrm{types}} = \{ v \in \mathcal{V} \mid \Psi(v) \in \mathrm{types} \}.
\]
For instance, \(\mathcal{V}_{\{N, R, S\}}\) represents the subset containing only entity, relationship, and semantic unit nodes.

\(\mathcal{V}_{\{T, S, A, H\}}\) contain rich informational content and are classified as retrievable nodes. In contrast, \(\mathcal{V}_{\{N, O\}}\), which represent names or titles, act solely as critical linkage and entry points within the graph but are not directly retrievable. For example, \(\mathcal{V}_{H}\) provides detailed context for high-level concepts, while \(\mathcal{V}_{O}\) represents the corresponding title and keywords but does not contribute directly to the retrieved content. Additionally, \(\mathcal{V}_{R}\), is a nodalized edge, acting as connector nodes and secondary retrievable nodes, contributing to the retrieval context but not serving as graph entry points.

\subsection{Graph Decomposition}
\label{Decomposition}
First, we define a null heterograph \(\mathcal{G}^0\). The initial step involves employing a LLM to decompose text chunks from the source corpus into three primary node types: semantic units (\(S\)), entities (\(N\)), and relationships (\(R\)). These nodes are then interconnected to construct the initial heterograph. This process can be formalized as:  
\[
\mathcal{G}^1 = \mathcal{G}^0 \cup \{v \in \mathcal{V}, e_d, e_r \in \mathcal{E} \mid \Psi(v) \in \{S, N, R\}\},
\]  
Where \(e\) represents the connecting edges between semantic units and entity nodes, as well as between relationship nodes and their corresponding source and target entities. For instance, if \textit{``Hinton was awarded the Nobel Prize for inventing backpropagation''} serves as \(v \in \mathcal{V}_{S}\) derived from a text chunk, then \textit{Hinton}, \textit{Nobel Prize}, and \textit{backpropagation} represent \(v \in \mathcal{V}_{N}\) nodes, with \(e_d\) denoting their connections to \(v \in \mathcal{V}_{S}\). An example of \(v \in \mathcal{V}_{R}\) would be \textit{``Hinton received Nobel Prize''}, where \(e_r\) represents the edge connecting the source node \textit{Hinton} to the target node \textit{Nobel Prize}.

\paragraph{Semantic unit (\(S\))}The semantic unit acts as a local summary, representing an independent event unit in a paraphrased form. It serves as the core node for graph augmentation and improving search quality. Since the division of text chunks is not based on semantics, unrelated or unassociated content may coexist within a single chunk. This context noise increases entropy, leading to degraded quality when using text chunks for graph augmentation or searching due to their coarse granularity and irrelevant information.

\paragraph{Entity (\(N\)) and Relationship (\(R)\)} Entities (\(N\)) are nodes that exclusively represent entity names, while relationships (\(R\)) are also transformed into nodes that connect source and target entities. These entities and relationships are directly connected to semantic units (\(S\)), as \(v \in \mathcal{V}_{S}\) serves as the smallest, contextually meaningful representation of events within text chunks. This connection ensures that entities and relationships remain decoupled from specific events, allowing them to function independently while still being anchored to relevant contexts. Such a design prevents redundant information and enables a flexible graph structure. 

\subsection{Graph Augmentation}
\label{augmentation}
The heterograph \(\mathcal{G}^1\) provides a foundational low-level structure. However, it lacks high-level organization and contextual insights. To further augment the graph, we implement two primary methods: node importance-based augmentation and community detection-based aggregation, which respectively capture the perspectives of individual node significance and structural cohesion within the graph.

\paragraph{Node Importance Based Augmentation} 
We prioritize the selection of structurally significant and functionally pivotal entities. These key entities, along with their associated semantic units and relationships, are processed through LLMs to generate attribute summaries. This approach mirrors human reading behavior, where all relevant content associated with a critical entity is reviewed before synthesizing its attributes. The summarization specifically focuses on the important entities identified within the corpus, rather than processing all entities, ensuring both precision and efficiency. The selection of important entities, \(N^*\), is guided by two complementary metrics: \(K\)-core decomposition \cite{seidman1983network,kong2019k} and betweenness centrality \cite{brandes2001faster}. \(K\)-core identifies nodes in densely connected subgraphs that are critical to graph cohesion, while betweenness centrality highlights nodes that act as bridges for information flow. These metrics are denoted as \(K(\mathcal{G}^1)\) and \(B(\mathcal{G}^1)\), where \(K(\cdot)\) and \(B(\cdot)\) represent the selected entity nodes from the graph. The final set of important entities is defined as:
\[
N^* = K(\mathcal{G}^1) \cup B(\mathcal{G}^1).
\]

Entity attributes are constructed directly from relationships and semantic units, bypassing raw texts to avoid redundancy. Each generated attribute node is added to the graph and connected to its corresponding entity node via the edge \(e_a\). This update to the graph is represented as:  
\[
\mathcal{G}^2 = \mathcal{G}^1 \cup \{v \in \mathcal{V}, e_a \in \mathcal{E} \mid \Psi(v) \in \{A\}\}. 
\]

\paragraph{Community Detection Based Aggregation}
We first apply the Leiden algorithm \cite{traag2019louvain} to \(\mathcal{G}^2\) to perform community detection, segmenting the graph into closely related substructures, denoted as communities. Each node \(v \in \mathcal{G}^2\) is assigned to a specific community \(\mathcal{C}_n\), where \(\mathcal{C}_n\) represents the \(n\)-th community identified by the algorithm. Within each community \(\mathcal{C}_n\), an LLM is utilized to analyze the aggregated content, extracting high-level elements (\(H\)) that encapsulate the core information of the community, such as summaries, sentiment analysis, and other significant insights. For each generated high-level element node \(v \in \mathcal{V}_{H}\), it is essential to establish meaningful connections \(e_h\) with relevant nodes from \(\mathcal{G}^2\) to preserve the graph's structural coherence. To accomplish this, we propose semantic matching within community algorithm. This algorithm identifies the most semantically related nodes within the same community \( \mathcal{C}_n \) for each high-level element node. To achieve this, K-means clustering \cite{macqueen1967some} is applied to the embeddings of \(v \in \mathcal{V}_{\{S,A,H\}}\). The number of clusters \( K \) is determined as \( \sqrt{ |\mathcal{V}_{\{S, A, H\}}| }\), where \( |\mathcal{V}_{\{S, A, H\}}| \) represents the total number of nodes labeled \( S \), \( A \), or \( H \). An edge \( e_h(v, v') \) exists between \( v \in \mathcal{V}_{\{S,A,H\}} \) and \( v' \in \mathcal{V}_{H} \) if both \( v \) and \( v' \) belong to the same semantic cluster \( \mathcal{S}_k \) and the same community \( \mathcal{C}_n \). Additionally, the LLM can extract a keyword title for each high-level element, referred to the high-level overview (\(O\)), which is used for dual search as elaborated in Section \ref{searching}. Each \(v \in \mathcal{V}_{H}\) and \(v \in \mathcal{V}_{O}\) will have a corresponding connection \(e_o\). The updated graph \(\mathcal{G}^3\) incorporates high-level elements (\(H\)) and their corresponding connections (\(e_h, e_o\)). It is defined as:
\[
\mathcal{G}^3 = \mathcal{G}^2 \cup \{v \in \mathcal{V}, e_h, e_o \in \mathcal{E} \mid \psi(v) = \{H, O\}\}.
\]
\subsection{Graph Enrichment}
\label{enrichment}
In the previous process of generating the heterograph, \(\mathcal{G}^3\) already contains a wealth of information. However, certain unique and additional details can still further enrich the heterograph, enabling it to not only preserve the entirety of the original text's information but also gain enhanced features and insights that go far beyond the source material.

\paragraph{Text Insertion} As mentioned earlier, text chunks are not directly incorporated into \(\mathcal{G}\) during graph augmentation due to their semantic incoherent nature. However, original text chunks hold significant value as they contain detailed information, which is often lost during the LLM transformation process. Therefore, it is essential to ensure that the original information remains searchable within graph. 
\[
\mathcal{G}^4 = \mathcal{G}^3 \cup \{v, e_s \mid \Psi(v) = T\},
\]
where \(e_s\) denotes the edges connecting text chunks to their relevant semantic units. 

\paragraph{Embedding}

As mentioned in Section \ref{NodeGraph}, \(v \in \mathcal{V}_{\{T,A,S,H\}}\) contains rich informational context where vector similarity is highly effective. Conversely, \(v \in \mathcal{V}_{\{N,O\}}\), which includes names and titles represented as words or phrase, is less suitable for vector similarity methods. To address this limitation, we developed a dual search mechanism. During the embedding process, we selectively embed only a subset of the graph's data, specifically \(v \in \mathcal{V}_{\{T,A,S,H\}}\). This targeted embedding step is crucial for reducing storage overhead while preserving efficient search capabilities.

\paragraph{HNSW Semantic Edges} The Hierarchical Navigable Small World (HNSW) algorithm \cite{malkov2018efficient} is an approximate nearest neighbor search method that organizes data into a multi-layer graph structure to efficiently retrieve semantically similar nodes. It represents the data as a layered graph \(\mathcal{H} = \{\mathcal{L}_0, \mathcal{L}_1, \dots, \mathcal{L}_m\}\), where \(\mathcal{L}_0\) is the base layer containing the densest semantic similarity connections, and higher layers (\(\mathcal{L}_i, i > 0\)) are sparsely connected to facilitate coarse-grained navigation. \(\mathcal{H}\) is built iteratively. When a new node is added, it is inserted into a random level and all layers below it, connecting to similar neighbors based on cosine similarity. Higher layers remain sparse with long-range connections, while \(\mathcal{L}_0\) focuses on dense local relationships. The search starts at the sparsely connected top layer, and progressively descends to \(\mathcal{L}_0\). In our work, the base layer \(\mathcal{L}_0\) of the HNSW graph, which encodes semantic relations between nodes, is integrated with the heterograph \(\mathcal{G}\). The updated graph, denoted as \(\mathcal{G}^5\), is expressed as:  
\[
\mathcal{G}^5 = \mathcal{G}^4 \cup \mathcal{L}_0.
\]  
The inclusion of \(\mathcal{L}_0\) enhances the heterograph's search capabilities by incorporating semantic dense proximity edges, augmenting its structural information in the graph. When an edge already exists in \(\mathcal{G}^4\), adding the corresponding edge from \(\mathcal{L}_0\) increases its weight by 1, reinforcing frequently occurring connections.

\begin{figure*}[ht]
    \centering
    \includegraphics[width=\textwidth]{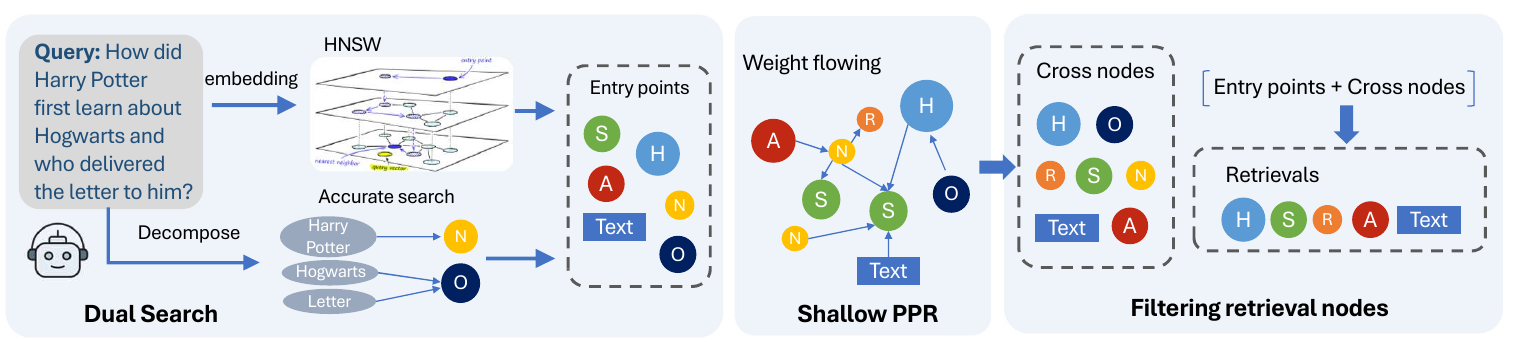} 
    \setlength{\abovecaptionskip}{0.1cm}
    \caption{This figure focuses on the querying process, where entry points are extracted from the original query, followed by searching for related nodes that need to be retrieved in the heterograph.}
    \label{fig:process2}
\end{figure*}

\subsection{Graph Searching}
\label{searching}

We first apply a dual search mechanism to identify entry points within the heterograph. Subsequently, a shallow Personalized PageRank (PPR) algorithm is employed to extract cross nodes. The combination of entry point nodes and cross nodes is then filtered to produce the final retrieval.

\paragraph{Dual Search}
Dual search combines exact matching on title nodes and vector similarity search on rich information nodes to identify entry points in the heterograph \(\mathcal{G}\). Given a query, the LLM extracts entities \(N^q\) and embeds the query into vector (\(\mathbf{q}\)). The entry points are defined as:
\[
\mathcal{V}_{\text{entry}} = \{v \in \mathcal{V} \mid \Phi(v, N^q, \mathbf{q})\},
\]
where the condition function \(\Phi(v, N^q, \mathbf{q})\) is defined as:
\[
\Phi(v, N^q, \mathbf{q}) = 
\begin{cases} 
v \in \mathcal{V}_{\{N, O\}} \land \mathcal{M}(N^q, v), \\ 
v \in \mathcal{V}_{\{S, A, H\}} \land \mathcal{R}(\mathbf{q}, v, k).
\end{cases}
\]

Here, the exact matching function \(\mathcal{M}(v^*, v)\) returns true if a node matches one of the extracted entities by word level string matching. Additionally, the similarity-ranking function \(\mathcal{R}(\mathbf{q}, v, k)\) returns true if a node ranks among the top-\(k\) most similar to \(\mathbf{q}\) based on the HNSW algorithm. By leveraging the non-retrievability of  \(v \in \mathcal{V}_{\{N,O\}}\), they serve exclusively as entry points to the graph without contributing directly to the retrievable content. Only nodes identified through the shallow PPR as closely related to all entry points are included in the retrieval results as cross nodes. This ensures that the effects of noisy or ambiguous queries, which may lead to errors in exact matching, do not directly impact the retrieval process. Any indirect effects are further minimized by the graph algorithm, enhancing the robustness of the retrieval system. 

\begin{table*}[!ht]
\centering
\scriptsize
\setlength{\tabcolsep}{2pt} 
\renewcommand{\arraystretch}{1.4} 

\textbf{Part I: General comparisons} \\[1ex]
\resizebox{\textwidth}{!}{
\begin{tabular*}{1.2\textwidth}{@{\extracolsep{\fill}}lcccccc|cccccccccccc@{}}
\toprule
\multirow{2}{*}{\textbf{Methods}} 
& \multicolumn{2}{c}{\textbf{HotpotQA}} 
& \multicolumn{2}{c}{\textbf{MuSiQue}} 
& \multicolumn{2}{c}{\textbf{MultiHop}} 
& \multicolumn{2}{c}{\textbf{Arena-Writing}} 
& \multicolumn{2}{c}{\textbf{Arena-Tech}} 
& \multicolumn{2}{c}{\textbf{Arena-Science}} 
& \multicolumn{2}{c}{\textbf{Arena-Recreation}} 
& \multicolumn{2}{c}{\textbf{Arena-Lifestyle}}
& \multicolumn{2}{c}{\textbf{Arena-FiQA}} \\
\cmidrule(lr){2-3} \cmidrule(lr){4-5} 
\cmidrule(lr){6-7} 
\cmidrule(lr){8-9} \cmidrule(lr){10-11} \cmidrule(lr){12-13} 
\cmidrule(lr){14-15} \cmidrule(lr){16-17} \cmidrule(lr){18-19}
& \textbf{Acc.$\uparrow$} & \textbf{\#Token$\downarrow$} 
& \textbf{Acc.$\uparrow$} & \textbf{\#Token$\downarrow$} 
& \textbf{Sco.$\uparrow$} & \textbf{\#Token$\downarrow$} 
& \textbf{W+T$\uparrow$} & \textbf{\#Token$\downarrow$} 
& \textbf{W+T$\uparrow$} & \textbf{\#Token$\downarrow$} 
& \textbf{W+T$\uparrow$} & \textbf{\#Token$\downarrow$} 
& \textbf{W+T$\uparrow$} & \textbf{\#Token$\downarrow$} 
& \textbf{W+T$\uparrow$} & \textbf{\#Token$\downarrow$} 
& \textbf{W+T$\uparrow$} & \textbf{\#Token$\downarrow$} \\
\midrule
NaiveRAG 
& 87.50\% & 9.8k 
& 39.43\% & 9.6k 
& 0.56 & 8.9k 
& 0.663 & 9.4k 
& 0.689 & 9.1k 
& 0.526 & 9.0k 
& 0.720 & 9.3k 
& 0.817 & 9.1k 
& 0.926 & 9.1k \\
HyDE     
& 73.00\% & 10.0k 
& 33.14\% & 9.8k 
& 0.53 & 9.4k 
& 0.789 & 9.6k 
& 0.863 & 9.3k 
& 0.823 & 9.3k 
& 0.777 & 9.5k 
& 0.829 & 9.3k 
& 0.949 & 9.3k \\
\midrule
LightRAG 
& 79.00\% & 7.1k 
& 36.00\% & 7.4k 
& 0.50 & 7.9k 
& 0.754 & 6.3k 
& 0.937 & 6.9k 
& 0.840 & 7.1k 
& 0.800 & 6.2k 
& 0.817 & 6.8k
& 0.937 & 7.7k \\
GraphRAG 
& 89.00\% & 6.6k 
& 41.71\% & 6.6k 
& 0.53 & 7.4k 
& 0.749 & 6.4k 
& 0.943 & 6.7k 
& 0.863 & 6.7k 
& 0.806 & 6.6k 
& 0.863 & 6.8k
& 0.960 & 6.8k \\
NodeRAG  
& \textbf{89.50}\% & \textbf{5.0k} 
& \textbf{46.29}\% & \textbf{5.9k} 
& \textbf{0.57} & \textbf{6.1k} 
& \textbf{0.794} & \textbf{3.3k} 
& \textbf{0.949} & \textbf{3.8k} 
& \textbf{0.903} & \textbf{4.2k} 
& \textbf{0.886} & \textbf{3.4k} 
& \textbf{0.949} & \textbf{3.3k} 
& \textbf{0.977} & \textbf{3.4k} \\
\bottomrule
\end{tabular*}
}

\vspace{1em} 

\hspace*{0.4cm}\textbf{Part II: Pairwise Comparisons} \\[1ex]
\resizebox{\textwidth}{!}{
\begin{tabular*}{1.4\textwidth}{@{\extracolsep{\fill}}llccc|clccc|clccc@{}}
\toprule
\textbf{Domain} & \multicolumn{1}{c}{\textbf{M1 vs M2}} & \textbf{Win (M1)} & \textbf{Tie} & \textbf{Win (M2)} 
& \textbf{Domain} & \multicolumn{1}{c}{\textbf{M1 vs M2}} & \textbf{Win (M1)} & \textbf{Tie} & \textbf{Win (M2)} 
& \textbf{Domain} & \multicolumn{1}{c}{\textbf{M1 vs M2}} & \textbf{Win (M1)} & \textbf{Tie} & \textbf{Win (M2)} \\
\midrule
\multirow{10}{*}{\textbf{FiQA}} 
& NodeRAG vs GraphRAG & \textcolor{Green}{\textbf{0.520}} & 0.126 & 0.354 
& \multirow{10}{*}{\textbf{Recreation}} 
& NodeRAG vs GraphRAG & \textcolor{Green}{\textbf{0.531}} & 0.126 & 0.343 
& \multirow{10}{*}{\textbf{Writing}} 
& NodeRAG vs GraphRAG & \textcolor{Green}{\textbf{0.691}} & 0.120 & 0.189 \\
& NodeRAG vs LightRAG & \textcolor{Green}{\textbf{0.486}} & 0.103 & 0.411 
& & NodeRAG vs LightRAG & \textcolor{Green}{\textbf{0.526}} & 0.143 & 0.331 
& & NodeRAG vs LightRAG & \textcolor{Green}{\textbf{0.651}} & 0.115 & 0.234 \\
& NodeRAG vs NaiveRAG & \textcolor{Green}{\textbf{0.749}} & 0.034 & 0.217 
& & NodeRAG vs NaiveRAG & \textcolor{Green}{\textbf{0.800}} & 0.017 & 0.183 
& & NodeRAG vs NaiveRAG & \textcolor{Green}{\textbf{0.851}} & 0.018 & 0.131 \\
& NodeRAG vs HyDE & \textcolor{Green}{\textbf{0.531}} & 0.155 & 0.314 
& & NodeRAG vs HyDE & \textcolor{Green}{\textbf{0.440}} & 0.189 & 0.371 
& & NodeRAG vs HyDE & 0.349 & 0.228 & \textcolor{Green}{\textbf{0.423}} \\
& GraphRAG vs LightRAG & 0.320 & 0.303 & \textcolor{Red}{\textbf{0.377}} 
& & GraphRAG vs LightRAG & 0.406 & 0.154 & \textcolor{Red}{\textbf{0.440}} 
& & GraphRAG vs LightRAG & 0.297 & 0.303 & \textcolor{Red}{\textbf{0.400}} \\
& GraphRAG vs NaiveRAG & \textcolor{Red}{\textbf{0.754}} & 0.092 & 0.154 
& & GraphRAG vs NaiveRAG & \textcolor{Red}{\textbf{0.714}} & 0.080 & 0.206 
& & GraphRAG vs NaiveRAG & \textcolor{Red}{\textbf{0.691}} & 0.092 & 0.217 \\
& GraphRAG vs HyDE & \textcolor{Red}{\textbf{0.491}} & 0.132 & 0.377 
& & GraphRAG vs HyDE & 0.377 & 0.137 & \textcolor{Red}{\textbf{0.486}}
& & GraphRAG vs HyDE & 0.177 & 0.126 & \textcolor{Red}{\textbf{0.697}} \\
& LightRAG vs NaiveRAG & \textcolor{Red}{\textbf{0.711}} & 0.106 & 0.183 
& & LightRAG vs NaiveRAG & \textcolor{Red}{\textbf{0.691}} & 0.063 & 0.246 
& & LightRAG vs NaiveRAG & \textcolor{Red}{\textbf{0.731}} & 0.080 & 0.189 \\
& LightRAG vs HyDE & \textcolor{Red}{\textbf{0.514}} & 0.143 & 0.343 
& & LightRAG vs HyDE & 0.349 & 0.171 & \textcolor{Red}{\textbf{0.480}} 
& & LightRAG vs HyDE & 0.211 & 0.178 & \textcolor{Red}{\textbf{0.611}} \\
& NaiveRAG vs HyDE & \textcolor{Red}{\textbf{0.611}} & 0.063 & 0.326 
& & NaiveRAG vs HyDE & \textcolor{Red}{\textbf{0.674}} & 0.069 & 0.257 
& & HyDE vs NaiveRAG & \textcolor{Red}{\textbf{0.857}} & 0.040 & 0.103 \\
\midrule
\multirow{10}{*}{\textbf{Lifestyle}} 
& NodeRAG vs GraphRAG & \textcolor{Green}{\textbf{0.640}} & 0.114 & 0.246 
& \multirow{10}{*}{\textbf{Science}} 
& NodeRAG vs GraphRAG & \textcolor{Green}{\textbf{0.497}} & 0.200 & 0.303 
& \multirow{10}{*}{\textbf{Tech}} 
& NodeRAG vs GraphRAG & \textcolor{Green}{\textbf{0.543}} & 0.154 & 0.303 \\
& NodeRAG vs LightRAG & \textcolor{Green}{\textbf{0.623}} & 0.131 & 0.246 
& & NodeRAG vs LightRAG & \textcolor{Green}{\textbf{0.538}} & 0.208 & 0.254 
& & NodeRAG vs LightRAG & \textcolor{Green}{\textbf{0.497}} & 0.137 & 0.366 \\
& NodeRAG vs NaiveRAG & \textcolor{Green}{\textbf{0.800}} & 0.040 & 0.160 
& & NodeRAG vs NaiveRAG & \textcolor{Green}{\textbf{0.829}} & 0.085 & 0.086 
& & NodeRAG vs NaiveRAG & \textcolor{Green}{\textbf{0.777}} & 0.046 & 0.177 \\
& NodeRAG vs HyDE & \textcolor{Green}{\textbf{0.526}} & 0.205 & 0.269 
& & NodeRAG vs HyDE & \textcolor{Green}{\textbf{0.423}} & 0.280 & 0.297 
& & NodeRAG vs HyDE & \textcolor{Green}{\textbf{0.543}} & 0.160 & 0.297 \\
& GraphRAG vs LightRAG & 0.429 & 0.120 & \textcolor{Red}{\textbf{0.451}}
& & GraphRAG vs LightRAG & \textcolor{Red}{\textbf{0.361}} & 0.343 & 0.296 
& & GraphRAG vs LightRAG & \textcolor{Red}{\textbf{0.400}} & 0.234 & 0.366 \\
& GraphRAG vs NaiveRAG & \textcolor{Red}{\textbf{0.680}} & 0.074 & 0.246 
& & GraphRAG vs NaiveRAG & \textcolor{Red}{\textbf{0.829}} & 0.108 & 0.063 
& & GraphRAG vs NaiveRAG & \textcolor{Red}{\textbf{0.657}} & 0.097 & 0.246 \\
& GraphRAG vs HyDE & 0.354 & 0.097 & \textcolor{Red}{\textbf{0.549}} 
& & GraphRAG vs HyDE & 0.354 & 0.172 & \textcolor{Red}{\textbf{0.474}} 
& & GraphRAG vs HyDE & \textcolor{Red}{\textbf{0.463}} & 0.143 & 0.394 \\
& LightRAG vs NaiveRAG & \textcolor{Red}{\textbf{0.663}} & 0.046 & 0.291 
& & LightRAG vs NaiveRAG & \textcolor{Red}{\textbf{0.828}} & 0.119 & 0.053 
& & LightRAG vs NaiveRAG & \textcolor{Red}{\textbf{0.691}} & 0.075 & 0.234 \\
& LightRAG vs HyDE & 0.349 & 0.120 & \textcolor{Red}{\textbf{0.531}} 
& & LightRAG vs HyDE & 0.308 & 0.189 & \textcolor{Red}{\textbf{0.503}} 
& & LightRAG vs HyDE & \textcolor{Red}{\textbf{0.463}} & 0.097 & 0.440 \\
& HyDE vs NaiveRAG & \textcolor{Red}{\textbf{0.709}} & 0.028 & 0.263 
& & HyDE vs NaiveRAG & \textcolor{Red}{\textbf{0.840}} & 0.074 & 0.086 
& & HyDE vs NaiveRAG & \textcolor{Red}{\textbf{0.606}} & 0.051 & 0.343 \\
\bottomrule
\end{tabular*}}

\setlength{\abovecaptionskip}{0.3cm}

\caption{\textbf{Part I: General Comparisons} evaluates NaiveRAG, HyDE, LightRAG, GraphRAG, and NodeRAG on HotpotQA and MuSiQue (accuracy and average tokens) and in the Arena using \textit{Win+Tie ratios} and \textit{average tokens}. 
\textbf{Part II: Pairwise Comparisons} shows the fraction of ``wins" (Win(M1)), ``ties" (Tie), and ``losses" (Win(M2)) when comparing one RAG method against another (e.g., NodeRAG vs. GraphRAG). Bold values highlight the best performance.
}

\label{tab:rag_comparison}
\end{table*}

\paragraph{Shallow PPR}
Personalized PageRank (PPR) identifies relevant nodes in the heterograph \(\mathcal{G}\) by simulating a biased random walk starting from a set of entry points. In our approach, we use shallow PPR, limiting the number of iterations \(t\) to ensure that relevance remains localized to the neighborhoods of the entry points. This early stop strategy prevents excessive diffusion to distant or irrelevant parts of the graph, focusing instead on multi-hop nodes near the entry points. Let \(P\) be the normalized adjacency matrix of \(\mathcal{G}\), where \(P_{ij}\) represents the transition probability from node \(i\) to node \(j\). The PPR process starts with a personalization vector \(p \in \mathbb{R}^{|\mathcal{V}|}\), where \(p_i = 1/|\mathcal{V}_{\text{entry}}|\) if \(v_i \in \mathcal{V}_{\text{entry}}\), and \(p_i = 0\) otherwise. The PPR score vector \(\pi^{(t)}\) after \(t\) iterations is computed iteratively as:
\[
\pi^{(t)} = \alpha p + (1 - \alpha) P^\top \pi^{(t-1)}, \quad \pi^{(0)} = p,
\]
where \(\alpha \in (0, 1)\) is the teleport probability that balances restarting at entry points and propagating through the graph. After \(t\) iterations, the top-\(k\) nodes with the highest PPR scores for each type are selected as cross nodes, denoted as \(V_{\text{cross}}\). In our default setting, we use \(\alpha = 0.5\) and \(t = 2\) to achieve a balance between exploration and convergence.

\paragraph{Filter Retrieval Nodes} Finally, the retrieval nodes are filtered from the union of entry nodes and cross nodes to include only retrievable nodes of  \(v \in \mathcal{V}_{\{T,A,S,H,R\}}\).  \(v \in \mathcal{V}_{\{N,O\}}\), which contain only keywords without informational content, are excluded from the retrieval context. The final set of retrieval nodes is therefore defined as:  
\[
\begin{aligned}
V_{\text{retrieval}}
&= \{\,v \in \mathcal{V}_{\text{entry}} \cup \mathcal{V}_{\text{cross}} \mid \\
&\quad \quad \psi(v) \in \{T, S, A, H, R\}\}
\end{aligned}
\]

\section{Evaluation}
We evaluate NodeRAG's performance across three different multihop benchmarks, \textbf{HotpotQA} \cite{yang2018hotpotqa}, \textbf{MuSiQue} \cite{trivedi2022musique}, \textbf{MultiHop-RAG} \cite{tang2024multihop}, and an open-ended head to head evaluation \textbf{RAG-QA Arena} \cite{han2024rag} across six domains. And we compare our method against several strong and widely used RAG methods as baseline models, including \textbf{NaiveRAG} \cite{lewis2020retrieval}, \textbf{HyDE} \cite{gao2022precise}, \textbf{GraphRAG} \cite{edge2024local}, \textbf{LightRAG} \cite{guo2024lightrag}. The details of these datasets and baseline models are introduced in Appendix \ref{experiment details}.

\subsection{Metrics}

\paragraph{General Comparison}
In the first part, we evaluate NaiveRAG, HyDE, LightRAG, GraphRAG, and NodeRAG across four benchmark datasets. For HotpotQA and MuSiQue benchmarks, we assess accuracy (Acc) to measure effectiveness and the average number of retrieved tokens (\#Token) to evaluate efficiency. For the MultiHop-RAG benchmark, we adopt its original evaluation metric, Score (Sco), while still using \#Token to gauge retrieval efficiency. Lastly, for the RAG-QA Arena benchmark, we continue to track \#Token for efficiency and employ a win and tie ratio (W+T) against gold responses as a measure of performance across different methods.

\paragraph{Pairwise Comparsion}
In this part, the evaluation focuses exclusively on the RAG-QA Arena benchmark, covering six domains: FiQA, Recreation, Writing, Lifestyle, Science, and Technology. We conduct comprehensive pairwise comparisons among all method combinations and calculate the corresponding win and tie rates for each matchup, thereby identifying the better RAG system.

\subsection{Implementation details}
By default, all these RAG methods are implemented with GPT 4o-mini, and the temperature is set to 0 across the entire evaluation. Meanwhile, we identify a potential unfairness in the current evaluation setup, evident in several key areas. Notably, the baselines vary in their choice of prompts used to synthesis the final response based on retrieved information. Therefore, we standardized response prompts for every method. Our initiative to standardize these settings also benefits other methods like GraphRAG, improving their performance compared to their default setting, underscoring the broader value of establishing fair and consistent evaluation standards.

\subsection{Results}
\paragraph{General Comparison} 
As shown in Part I of Table \ref{tab:rag_comparison}, NodeRAG consistently outperforms competing methods on HotpotQA, MuSiQue, and MultiHopRAG, demonstrating the highest accuracy while retrieving noticeably fewer tokens. For example, for MuSiQue, NodeRAG attains an accuracy of $46.29\%$, surpassing GraphRAG ($41.71\%$) and LightRAG ($36.00\%$). In HotpotQA, while NodeRAG achieves a slightly higher accuracy ($89.50\%$ vs. $89.00\%$ for GraphRAG), it does so with only 5k retrieved tokens, which is 1.6k fewer than GraphRAG. In the RAG-QA Arena benchmark, graph-enhanced RAG systems exhibit a clear advantage over traditional approaches. Notably, NodeRAG achieves the highest win and tie ratio in each of the five domains while keeping retrieval costs minimal. For example, it attains a ratio of $94.9\%$, notably surpassing GraphRAG’s $86.3\%$ and LightRAG’s $81.7\%$ in the Lifestyle domain, and does so with less than half the retrieved tokens compared to the other models. It can also be noticed that graph-enhanced RAG systems generally retrieve fewer tokens than traditional RAG across all benchmarks. These results confirm NodeRAG’s remarkable effectiveness and efficiency, demonstrating that our heterograph can significantly boost RAG performance across diverse tasks.

\paragraph{Pairwise Comparsion}
Across all the six domains, NodeRAG consistently achieves higher win ratios against GraphRAG, LightRAG, NaiveRAG, and HyDE, demonstrating notable dominance, for instance, in the Lifestyle domain, NodeRAG achieves 0.640 win rate against GraphRAG, 0.623 against LightRAG, 0.800 against NaiveRAG and 0.526 against HyDE. GraphRAG, LightRAG, NaiveRAG, and HyDE show scattered successes, such as LightRAG edging out NaiveRAG (0.649 vs. 0.246) in Recreation, GraphRAG beats LightRAG (0.361 vs. 0.296) in Science, yet their overall win rates remain lower when compared to NodeRAG. Notably, these trends persist across other domains like Writing, Recreation, Science, and Tech, further underscoring NodeRAG’s leading position, followed by LightRAG and GraphRAG, showing the superiority of our method. 

In general, NodeRAG not only achieves the highest accuracy rate and the lowest retrieval token count in general benchmarks but also outperforms all other baselines in preference evaluation comparisons. This unparalleled performance in both accuracy and computational efficiency makes NodeRAG the optimal choice for a wide range of RAG tasks, from research applications to deployments in resource-constrained environments.






\section{Ablation experiments}
\label{ablations}

\begin{figure}[ht]
    \centering
    \includegraphics[width=\columnwidth]{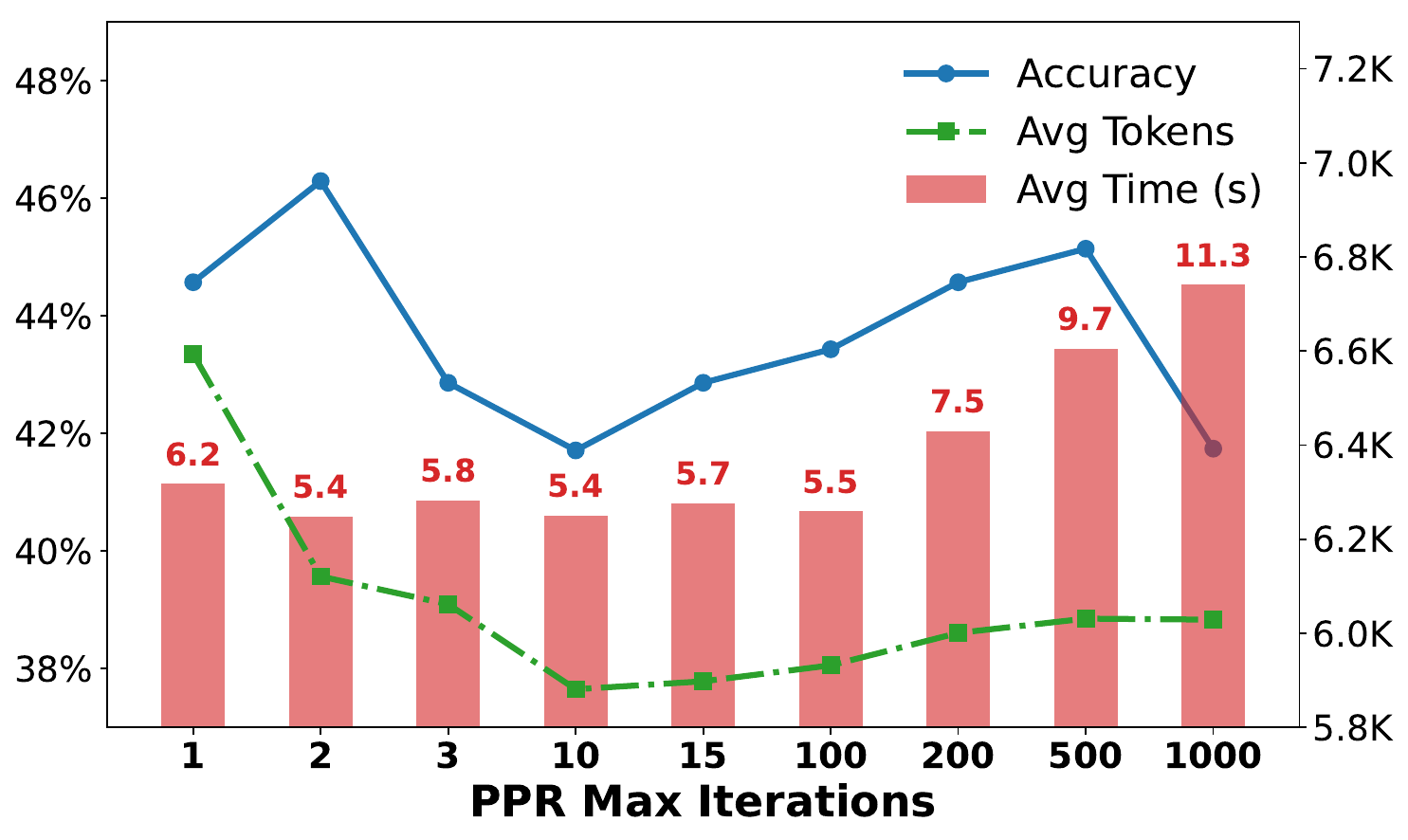} 
    \setlength{\abovecaptionskip}{-10pt}
    \caption{Ablation analysis on PPR iterations. }
    \label{fig:max_ppr}
\end{figure}

We conducted ablation experiments on the MuSiQue dataset, adhering to the same settings and evaluation metrics described earlier. We specifically examined the impact of four key submodules: shallow PPR, cross-node interactions, HNSW semantic edges, and dual search.

We first investigated the variation in PPR iterations and examined whether shallow PPR offers advantages. PPR, with a few iterations, performs better than deep PPR because it highlights important nodes that are closer to the entry points. Moreover, early stopping reduces unnecessary computational overhead, leading to improved retrieval efficiency.

Moreover, we evaluate the performance of applying the top-k vector similarity method to all node data in the graph. Although increasing the retrieval context, its performance remains lower than the basic version. This confirms the necessity of cross-nodes in our method, as they help identify important multi-hop nodes. Second, performing vector similarity solely on node data consistently outperforms the naive RAG approach of similarity on text chunks, demonstrating the advantages brought by graph-based data augmentation.

In addition, without integration of accurate search in dual search, accuracy drops to 44.57\%, and the token count increases to 9.7k. This is because losing entity and high-level overview nodes as entry points causes nodes with long texts, such as text nodes, to have higher weights after shallow PPR. Since vector similarity entry nodes are more frequently connected to $T$ nodes, while accurate entry nodes are more connected to $S$, $A$, and $H$ nodes, the absence of accurate search disrupts this balance.

Finally, we investigate the effect of HNSW. HNSW introduces semantic edges to the heterograph, and removing this integration results in performance degradation. This is because HNSW enhances connectivity between semantically related nodes, enabling more efficient and meaningful retrieval.

\begin{table}[htbp]
\centering
\small
\setlength{\tabcolsep}{6pt}
\renewcommand{\arraystretch}{1.1}
\begin{tabular}{lccc}
\toprule
\textbf{Method} & \textbf{Accuracy} & \textbf{Time (s)} & \textbf{Tokens (k)} \\
NodeRAG (Ours)      & 46.29\%          & 4.05             & 5.96               \\
\midrule
w/o HNSW           & 41.71\%          & 4.92             & 6.78               \\
w/o Dual Search    & 44.57\%          & 4.72             & 9.70               \\
\multicolumn{4}{l}{w/o Cross Node} \\
\hspace{4mm} Top-$k=10$ & 41.71\%      & 4.15             & 4.27               \\
\hspace{4mm} Top-$k=20$ & 43.43\%      & 4.70             & 7.89               \\
\hspace{4mm} Top-$k=30$ & 42.29\%      & 4.80             & 11.62              \\
\bottomrule
\end{tabular}
\setlength{\abovecaptionskip}{0.2cm}
\caption{Ablation study of NodeRAG components.}
\label{tab:ablation}
\end{table}

\section{Related Works}

\paragraph{Retrieval-augmented generation}
Retrieval-Augmented Generation (RAG) systems \cite{gupta2024comprehensive} enhance the performance of large language models (LLM) by retrieving relevant information from external documents, grounding responses in domain-specific knowledge. Traditional RAG approaches \cite{zhao2024retrieval} embed user queries and entries from a knowledge base into a shared vector space and then compare query vectors to knowledge base vectors to retrieve the top-$K$ most similar contexts based on cosine similarity or similar variants \cite{fan2024survey, lewis2020retrieval}. While effective, naive RAG methods face several limitations, prompting various enhancements in subsequent works. JPR \cite{min2021joint} improves multi-answer retrieval by refining passage selection, while IR-CoT \cite{trivedi2022interleaving} integrates chain-of-thought reasoning for multi-hop question answering. Similarly, Tree of Clarifications \cite{kim2023tree} constructs a tree-based disambiguation structure to resolve ambiguous queries. HyDE \cite{hyde} also enhances the performance of dense retrieval by generating hypothetical documents. Other works examine how different document types influence RAG effectiveness and LLM performance \cite{hsia2024ragged}. Despite these advancements, traditional RAG systems still face significant challenges. The context window limitations\cite{cheng2024lift, su2024dragin} of LLMs constrain their ability to process extensive external documents holistically \cite{jiang2024longrag}. RAG has been applied to various domain-specific knowledge bases, such as BioRAG and MedicalRAG \cite{wang2024biorag, wu2024medical, jiang2024tc}. RAG also struggles with corpus-wide understanding tasks, like query-focused abstractive summarization, which require synthesizing knowledge across large datasets.

\paragraph{RAG over Hierarchical Index}
To overcome the limitations of traditional RAG, advanced systems integrate hierarchical indexing to incorporate document summaries and enhance retrieval performance. Dense Hierarchical Retrieval (DHR) \cite{liu2021dense} improves passage representations by combining macroscopic document semantics with microscopic passage details. Expanding on this, Hybrid Hierarchical Retrieval (HHR) \cite{arivazhagan2023hybrid} fuses sparse and dense retrieval techniques for both document- and passage-level retrieval, achieving greater precision. Other methods leverage hierarchical data structures to facilitate complex document summarization. For instance, RAPTOR \cite{sarthi2024raptor} employs tree-based structures to integrate knowledge across lengthy documents, synthesizing information at various levels of abstraction. Graph-based RAGs \cite{trajanoska2023enhancing, zhang2024causal} extend this by constructing knowledge graphs (KGs) \cite{chen2020knowledge} at the indexing stage and applying graph algorithms during querying \cite{haveliwala2003analytical}. Notable examples include HippoRAG \cite{gutierrez2024hipporag} and KAPING \cite{baek2023knowledge}, which refine knowledge organization and retrieval efficiency. Similarly, GraphRAG \cite{edge2024local} introduces graph-based text indexing using LLMs and generates community-based summaries \cite{blondel2008fast, traag2019louvain}, inspiring subsequent works such as LightRAG \cite{guo2024lightrag}, which integrates both high- and low-level information while optimizing indexing costs. While these approaches effectively leverage hierarchical data structures, they do not fully exploit the synergy between LLMs and graph-based methods. Our proposed framework addresses these gaps by refining graph structures and incorporating advanced graph algorithms, leading to superior retrieval accuracy and efficiency.

\section{Conclusion and Discussion}

In this paper, we introduce NodeRAG, a novel framework designed to enhance RAG performance by optimizing graph structures in indexing for more effective and fine-grained retrieval. NodeRAG constructs a well-defined heterograph with functionally distinct nodes, balancing fine-grained understanding with a global perspective of the knowledge corpus. Experimental results demonstrate that NodeRAG outperforms existing methods across multi-hop reasoning benchmarks and open-ended retrieval tasks. As the saying goes,\textit{``A strong foundation supports a higher structure"}. In the realm of graph-based RAG, the graph structure serves as this very foundation. The introduction of NodeRAG underscores the critical role of graph structures, encouraging a renewed emphasis on their design and optimization. 

\bibliography{custom}

\newpage
\appendix
\setlength{\parindent}{0pt}
\setlength{\parskip}{6pt}
\onecolumn
\section{Comparison of RAG System Performance}
\label{RAGComparison}

\begin{table*}[ht!]
\centering
\resizebox{\textwidth}{!}{ 
\begin{tabular}{lccccccccccccccc}
\toprule
\textbf{Datasets} & \textbf{Corpus Size} & \multicolumn{3}{c}{\textbf{Index Time}} & \multicolumn{3}{c}{\textbf{Storage Usage}} & \multicolumn{4}{c}{\textbf{Query Time}} & \multicolumn{4}{c}{\textbf{Average Retrieval Tokens}} \\
\cmidrule(lr){3-5} \cmidrule(lr){6-8} \cmidrule(lr){9-12} \cmidrule(lr){13-16}
 &  & Graph & Light & Node & Graph & Light & Node & Graph-L & Graph-G & Light & Node & Graph-L & Graph-G & Light & Node \\
\midrule
HotpotQA & 1.93M & 66min & 39min & 21min & 227MB & 461MB & 214MB & 2.66s & 26.69s & 5.58s & 3.98s & 6680.65 & 810529 & 7176.73 & 5079.40 \\
Musique & 1.84M & 76min & 90min & 25min & 255MB & 492MB & 250MB & 2.94s & 22.65s & 6.53s & 4.05s & 6616.84 & 1111073 & 7458.34 & 5960.25 \\
MultiHop & 1.41M & 50min & 58min & 24min & 141MB & 276MB & 137MB & 4.15s & 34.45s & 7.10s & 4.89s & 7367.54 & 920780 & 8920.00 & 5259.99 \\
Arena-Fiqa & 1.65M & 45min & 49min & 19min & 112MB & 240MB & 117MB & 8.95s & 28.94s & 13.35s & 8.86s & 6819.45 & 713560 & 7721.73 & 3381.72 \\
Arena-Lifestyle & 1.64M & 52min & 59min & 18min & 138MB & 278MB & 125MB & 7.54s & 33.09s & 10.43s & 6.79s & 6860.26 & 895964 & 6822.32 & 3350.35 \\
Arena-Recreation & 0.93M & 34min & 33min & 10min & 89MB & 172MB & 80MB & 5.10s & 23.10s & 8.01s & 6.90s & 6669.95 & 564636 & 6249.31 & 3448.38 \\
Arena-Science & 1.43M & 43min & 46min & 17min & 116MB & 236MB & 111MB & 8.05s & 35.79s & 14.28s & 8.85s & 6759.15 & 778051 & 7111.80 & 4284.13 \\
Arena-Tech & 1.72M & 54min & 54min & 14min & 133MB & 276MB & 139MB & 7.35s & 28.64s & 8.89s & 6.74s & 6755.46 & 741690 & 6922.55 & 3821.78 \\
Arena-Writing & 1.82M & 50min & 71min & 13min & 151MB & 309MB & 157MB & 5.65s & 40.12s & 10.70s & 5.40s & 6477.72 & 877354 & 6364.59 & 3373.34 \\
\bottomrule
\end{tabular}
}
\setlength{\abovecaptionskip}{0.2cm}
\caption{Performance metrics for RAG methods, including Index Time, Storage Usage, Query Time, and Retrieval Tokens across various datasets. \textit{Graph} denotes GraphRAG, with \textit{Graph-l} representing its local mode and \textit{Graph-G} its global mode. \textit{Light} refers to LightRAG in hybrid mode, while \textit{Node} represents our proposed method.
}
\label{system performace}
\end{table*}

The table \ref{system performace} presents the system performance of mainstream graph-based RAG methods and our proposed approach. Compared to previous work, our method demonstrates superior performance across multiple datasets and in open-ended head-to-head evaluations, while also achieving better system-level efficiency. All evaluations in the table were conducted using the default indexing settings of each RAG method, with the query settings and the prompt details provided in Appendix \ref{baseline}. Notably, our method demonstrates a significant advantage in indexing time, which is crucial for practical deployment. This advantage is attributed to the construction process of our Hetero Graph, which not only creates a more fine-grained and semantically meaningful graph structure but also carefully considers the algorithmic complexity of the retrieval process.

NodeRAG also exhibits relatively better storage efficiency. Although the total number of nodes in our expanded graph is significantly larger than in previous graph structures, the combination of selective embedding and dual search effectively reduces the number of embedded nodes, leading to a more efficient storage strategy. Moreover, our unified information retrieval approach results in reduced query time. While the GraphRAG local search (Graph-l) relies purely on vector similarity—similar to our "without cross-node" setting mentioned in Section 4—and achieves faster search speeds, its global mode (Graph-G) experiences significantly higher query times, exceeding 20 seconds with a concurrency of 16. This is due to its reliance on LLM-based traversal of all community information, leading to a substantial number of retrieval tokens. Given the considerable time and computational overhead associated with Graph-G queries, we conducted a full evaluation only on the MuSiQue dataset. For other datasets, query time and retrieval token statistics were estimated based on a sample of 20 selected queries. Further details on the ablation study of GraphRAG can be found in the Appendix \ref{Graph RAG ablation}.

In contrast, our method leverages the heterograph and graph algorithms to achieve unified information retrieval, effectively capturing meaningful information needs across multiple levels within a single framework while maintaining efficient query speed. Finally, the nodes within the heterograph are connected in a fine-grained structure, ensuring that more relevant text is retrieved with relatively fewer retrieval tokens.

\section{Experiment details}
\label{experiment details}
\subsection{Datasets}

We evaluate Node RAG's performance across four different benchmarks: HotpotQA, MuSiQue, MultiHop-RAG and RAG-QA Arena. However, the original question formats of HotpotQA and MuSiQue required selecting the most relevant passages from multiple documents, incorporating multi-hop reasoning details. This setup no longer aligns with mainstream RAG methods, as modern approaches perform indexing over an entire corpus and subsequently retrieve information from the indexed data. To adapt to this paradigm, we concatenate all passages into a unified corpus, transforming the task into retrieving multi-hop relevant information from the entire corpus. This modification makes the task more challenging compared to the original setting. 
The evaluation metrics for HotpotQA and MuSiQue are divided into two aspects: the quality of the retrieved documents and the accuracy of the final answer, measured by metrics such as F1 score. However, current RAG methods retrieve not only text chunks but also more flexible forms of information, making it difficult to assess retrieval quality using traditional top-$k$ document evaluation. Moreover, metrics like F1 score have become less effective in evaluating answers generated by modern generative models. Therefore, we adopt the \textit{LLM-as-a-Judge} approach, leveraging LLMs to assess the final accuracy of the generated answers.The MultiHop and RAG-QA Arena dataset settings provide a strong evaluation framework for current RAG methods. Therefore, we follow the original benchmark's proposed testing methodology and evaluation metrics. Further details regarding the benchmark settings are described below.

\textbf{HotpotQA} is a multi-hop question-answering dataset where each question requires combining information from multiple documents to find the correct answer. It encourages deeper reasoning by providing supporting facts—specific sentences from the texts that lead to the solution. Questions range widely across domains and often involve bridging or comparison to ensure more complex, multi-step reasoning. This makes HotpotQA a critical benchmark for evaluating advanced reading comprehension models. We sampled 200 questions from the final dataset for evaluation.

\textbf{MuSiQue} is also a multi-hop question-answering dataset that challenges models to combine information across multiple documents in a structured, step-by-step manner. Each question is designed to require several reasoning steps, ensuring that simple “shortcut” approaches do not suffice. As a result, MuSiQue serves as a rigorous test of advanced reading comprehension, demanding that systems accurately connect disparate pieces of evidence to arrive at correct answers. We also sample 175 questions for the evaluation

\textbf{MultiHop-RAG} is a multi-hop question-answering dataset that includes four distinct question types: comparison query, null query, inference query, and temporal query. From this dataset, we curated 375 questions to evaluate our approach. Each query in MultiHop requires synthesizing information from multiple sources, testing a model’s ability to perform bridging inferences, handle temporal relationships, and make higher-order logical connections. This diversity in question types provides a rigorous benchmark for assessing whether RAG methods can integrate scattered pieces of evidence.

\textbf{RAG-QA Arena} is a new evaluation framework designed to assess the quality of retrieval-augmented generation (RAG) systems on long-form question answering. It builds on Long-form RobustQA (LFRQA), a dataset of 26K queries across seven domains including writing, tech, science, recreation and lifestyle. Each LFRQA entry features a coherent, human-written answer grounded in multiple documents. RAG-QA Arena leverages LLMs as evaluators, directly comparing a system's generated answer with the 'gold' long-form answer from LFRQA. Experimental results show that these model-based comparisons correlate highly with human judgments, making it a challenging yet reliable benchmark for testing both cross-domain robustness and the ability to produce integrated, long-form responses. 

\subsection{Baselines}

\label{baseline}
We compare NodeRAG against several strong and widely used RAG methods. By default, all these RAG methods implement their indexing process using GPT-4o-mini. However, we identify a potential unfairness in the current evaluation setup, particularly in several key areas.
To ensure the correctness and validity of the evaluation data, it is crucial to standardize both the final answer response prompt and the model temperature settings. Using different response prompts or varying temperature settings for answer generation introduces inconsistencies, as a higher temperature setting may yield responses that receive a better LLM preference score compared to those generated with a lower temperature. A critical point to consider is that, as RAG methods, the primary focus of evaluation should be the quality of the retrieved context rather than the final generated answer. Therefore, to ensure that final accuracy metrics accurately reflect the quality of the retrieved context, the final answer generation process and model settings should remain consistent across all methods. 
Hence, we set the temperature to 0 across the entire evaluation and standardized response prompts for every method. The unified prompt is illustrated in appendix \ref{Prompting in NodeRAG}. Our initiative to standardize these settings also benefits other methods, such as GraphRAG, improving their performance compared to their default settings. This underscores the broader value of establishing fair and consistent evaluation standards.
Additionally, traditional evaluation methods such as top-$k$ retrieval comparison have become increasingly difficult to apply uniformly, as retrieval is no longer restricted to isolated text chunks. To address this challenge, we propose a new evaluation standard that leverages retrieval tokens as an efficiency metric. This approach ensures that retrieval methods achieve better effectiveness while utilizing fewer retrieval tokens, promoting a more efficient and fair comparison framework. Current methods can only control the number of retrieval tokens through hyperparameter tuning. Although precise control over the exact number of tokens is not possible, we consider maintaining the average number of retrieval tokens within the range of 5K to 10K to be a reasonable and fair comparison criterion. Below, we provide a detailed introduction to each method along with its specific settings for reference.

\paragraph{Naive RAG}
This method serves as a standard baseline among all existing RAG systems. It first divided input document into several text chunks and encoded them into a vector space utilizing text embeddings. Then retrieve related text chunks based on similarity of query representations. The number of retrieval tokens can be adjusted through the top-$k$ parameter.

\paragraph{HyDE}
HyDE serves as an improved method over traditional RAG systems. It first generates "hypothetical" texts that capture the essence of a query. It then uses this generated text to retrieve relevant documents from a large corpus, employing vector similarity in an embedding space. This method modifies the input query at the frontend without altering the text chunks or their embeddings. Therefore, we can still use the top-$k$ parameter to control the number of retrieval tokens.

\paragraph{GraphRAG}
This approach starts by segmenting the input text into chunks and extracting the entities and relationships within them, forming a graph structure. This graph is then divided into multiple communities at different levels. At query time, GraphRAG identifies the relevant entities from the question and synthesizes answers by referencing these corresponding community summaries. Compared to traditional RAG methods, GraphRAG provides a more structured and high-level understanding of the entire document. Through our experiments, we observed that under the default settings, the number of queries in GraphRAG’s local mode resulted in a higher retrieval token count than the naive retrieval approach. To ensure a fair comparison, we proportionally reduced its parameters and standardized its prompt to match our unified prompt. The ablation study in Appendix \ref{Graph RAG ablation} demonstrates that after these adjustments, GraphRAG’s accuracy improved, further validating the fairness of our evaluation methodology. Additionally, we analyzed both the local and global modes of GraphRAG. Our findings indicate that the global mode introduces significant additional overhead in terms of time and computational cost while providing only marginal improvements compared to the local mode. This result is further supported by our ablation study, which shows that the local mode achieves better efficiency and effectiveness.

\paragraph{LightRAG}
LightRAG is an improved approach based on GraphRAG, designed to minimize computational overhead while enhancing the comprehensiveness of retrieved information through dual-level retrieval. This leads to more efficient retrieval and a better balance between effectiveness and speed compared to GraphRAG. Similar to GraphRAG, the default settings of LightRAG result in a higher retrieval token count than the Naïve approach. To ensure a fair comparison, we proportionally adjusted its hyperparameters to maintain the number of retrieval tokens within the range of 5K to 10K.

\subsection{NodeRAG Graph Statistics}

\begin{table*}[ht!]
\centering
\resizebox{\textwidth}{!}{ 
\begin{tabular}{lccccccccccccccc}
\toprule
\textbf{Datasets} & \textbf{Corpus Tokens} & \multicolumn{7}{c}{\textbf{Type Statistics}} & \multicolumn{4}{c}{\textbf{Graph Statistics}} \\
\cmidrule(lr){3-9} \cmidrule(lr){10-13}
 &  & \textbf{T} & \textbf{S} & \textbf{N} & \textbf{R} & \textbf{A} & \textbf{O} & \textbf{H} & \textbf{Nodes} & \textbf{Non-HNSW Edge} & \textbf{HNSW Edge} & \textbf{Edge} \\
\midrule
HotpotQA & 1.93M & 1985 & 15905 & 88863 & 56578 & 684 & 4479 & 4479 & 172603 & 283543 & 487731 & 759812 \\
MuSiQue & 1.84M & 1907 & 18714 & 99840 & 61964 & 795 & 5700 & 5700 & 193922 & 316029 & 583126 & 888966 \\
MultiHop-RAG & 1.41M & 1532 & 10986 & 43184 & 29286 & 685 & 2289 & 2289 & 90144 & 171410 & 203199 & 367486 \\
Arena-Fiqa & 1.65M & 1821 & 9027 & 32470 & 27422 & 508 & 1714 & 1714 & 74605 & 143916 & 154109 & 295165 \\
Arena-Lifestyle & 1.64M & 1794 & 9400 & 39464 & 27895 & 518 & 2221 & 2221 & 83461 & 149225 & 174461 & 318073 \\
Arena-Recreation & 0.93M & 1003 & 5542 & 26382 & 16938 & 413 & 1969 & 1969 & 54180 & 93228 & 117915 & 207449 \\
Arena-Science & 1.43M & 1583 & 8010 & 32232 & 23092 & 551 & 2515 & 2515 & 70425 & 127719 & 149424 & 276963 \\
Arena-Tech & 1.72M & 1910 & 10837 & 37724 & 29691 & 534 & 2633 & 2633 & 85888 & 167950 & 193159 & 354033 \\
Arena-Writing & 1.82M & 1937 & 11008 & 42723 & 29338 & 705 & 4435 & 4435 & 94259 & 149552 & 298565 & 442397 \\
\bottomrule
\end{tabular}
}
\setlength{\abovecaptionskip}{0.2cm}
\caption{Comprehensive dataset statistics, detailing corpus size, type statistics (T, S, N, R, A, O, H), and graph statistics. The graph statistics include the number of document compilation nodes, HNSW semantic edges, and total edges. Each value represents a key metric relevant to graph-based document processing and retrieval.}
\label{graph statistic}

\end{table*}

The table \ref{graph statistic} presents the number of each type of node in the indexed graph for each dataset, including entity (\(N\)), relationship (\(R\)), semantic unit (\(S\)), attribute (\(A\)), high-level elements (\(H\)), high-level overview (\(O\)), and text (\(T\)). These counts are detailed in the type statistics section. Additionally, the graph statistics provide information on the total number of nodes, the number of non-HNSW edges, HNSW edges, and the total number of edges. The data indicate that the number of HNSW edges is comparable to that of non-HNSW edges, highlighting the integration of semantic connections within the graph. Notably, overlapping edges are removed when merging non-HNSW and HNSW edges. For instance, in the MultiHop-RAG benchmark, there are 171,410 non-HNSW edges and 203,199 HNSW edges. However, the total number of edges after merging is 367,486, which is only 7,123 fewer than the sum of both edge types. This indicates the uniqueness of these two types of edges and highlights the effectiveness of the HNSW algorithm.

\subsection{Graph RAG Ablation}
\label{Graph RAG ablation}

\begin{table}[!htbp]
    \centering
    \label{tab:graphrag_performance}
    \resizebox{0.5\textwidth}{!}{
    \begin{tabular}{lccc}
        \toprule
        \textbf{Method} & \textbf{Accuracy} & \textbf{Avg. Processing Time} & \textbf{Avg. Tokens} \\
        \midrule
        GraphRAG (default) & 37.14\% & 4.82s & 10.4k \\
        Graph-L   & 41.71\% & 2.94s & 6.6k  \\
        Graph-G  & 33.14\% & 22.65s & 1.11M \\
        \bottomrule
    \end{tabular}
    }
    \setlength{\abovecaptionskip}{0.2cm}
    \caption{Performance Comparison of GraphRAG Variants. Default is the default setting. Local and global represent the local and global modes under unified prompt and hyperparameter settings.}
    
\end{table}
The default setting of GraphRAG, along with its own prompting mechanism, is not standardized for evaluation, as both the number of retrieval tokens and the choice of prompts significantly impact performance. Hence, we introduce a unified prompt and adjust the hyperparameters of GraphRAG to ensure a fair comparison within a specific range.  
As shown in the table \ref{Graph RAG ablation}, GraphRAG with our unified prompt achieves higher performance, demonstrating that the original prompting strategy is not optimal for this task. This further ensures fairness in comparison, as performance is influenced solely by the quality of the retrieved context.  
Moreover, the global mode of GraphRAG requires significantly longer processing time and incurs higher computational costs due to the LLM analyzing all community summaries, leading to increased complexity and resource consumption. Additionally, for multi-hop question answering, this approach results in degraded performance.  
Therefore, we conducted an exploratory ablation study only on the MuSiQue dataset, while for other datasets, we estimated query time and retrieval token statistics based on sampled queries.

\newpage
\section{Algorithm details}
\label{Algorithms}

\subsection{Terminology}
\label{Terminology}

\begin{table}[htbp]
\centering
\resizebox{\textwidth}{!}{
\begin{tabular}{|c|c|l|l|l|}
\hline
\textbf{Abbr.} & \textbf{Full Name} & \textbf{Description} & \textbf{Function} & \textbf{Example} \\ \hline

\(T\) 
& Text 
& \begin{minipage}{4.5cm} \vspace{3pt}
  Full-text chunks from the original source. It contains rich detailed information, although it integrates a large amount of unrelated semantic information.
  \vspace{3pt}
  \end{minipage}
& \begin{minipage}{3.5cm} \vspace{3pt}
  Retrievable; \\ Entry points from vector similarity
  \vspace{3pt} \end{minipage}
& \begin{minipage}{4.8cm} \vspace{3pt}
  \textit{"Hinton was awarded the Nobel Prize in 2023 for his groundbreaking contributions to artificial intelligence, particularly in deep learning. 
  His pioneering work on backpropagation laid the foundation for modern neural networks, influencing both academia and industry. 
  The recognition came amid increasing discussions on the ethical implications of AI, with Hinton himself advocating for responsible AI development and regulation."}
  \vspace{3pt} \end{minipage} \\ \hline

\(S\) 
& Semantic Unit 
& \begin{minipage}{4.5cm} \vspace{3pt}
  Local summaries that are independent and meaningful events summarized from text chunks. They serve as a middle layer between text chunks and entities, acting as the basic units for graph augmentation and semantic analysis.
  \vspace{3pt} \end{minipage}
& \begin{minipage}{3.5cm} \vspace{3pt}
  Retrievable; \\ Entry points from vector similarity.
  \vspace{3pt} \end{minipage}
& \begin{minipage}{4.8cm} \vspace{3pt}
  \textit{"Hinton was awarded the Nobel Prize for inventing backpropagation."}
  \vspace{3pt} \end{minipage} \\ \hline

\(A\) 
& Attribute 
& \begin{minipage}{4.5cm} \vspace{3pt}
  Attributes of key entities, derived from relationships and semantic units around important entities.
  \vspace{3pt} \end{minipage}
& \begin{minipage}{3.5cm} \vspace{3pt}
  Retrievable; \\ Entry points from vector similarity.
  \vspace{3pt} \end{minipage}
& \begin{minipage}{4.8cm} \vspace{3pt}
  \textit{"Geoffrey Hinton, often referred to as the "Godfather of Deep Learning," is a pioneer in the field of artificial intelligence. In 2024, he was awarded the Nobel Prize for his contributions to AI and deep learning. "}
  \vspace{3pt} \end{minipage} \\ \hline

\(H\) 
& High-Level Element 
& \begin{minipage}{4.5cm} \vspace{3pt}
  Insights summarizing graph communities. Encapsulates core information or any high level ideas from a community.
  \vspace{3pt} \end{minipage}
& \begin{minipage}{3.5cm} \vspace{3pt}
  Retrievable; \\ Entry points from vector similarity.
  \vspace{3pt} \end{minipage}
& \begin{minipage}{4.8cm} \vspace{3pt}
  \textit{"Due to the increasing importance of AI, the Nobel Prize is awarded to scholars who have made tremendous contributions to the field of AI."}
  \vspace{3pt} \end{minipage} \\ \hline

\(O\) 
& High-Level Overview 
& \begin{minipage}{4.5cm} \vspace{3pt}
  Titles or keywords summarizing \\ high-level elements.
  \vspace{3pt} \end{minipage}
& \begin{minipage}{3.5cm} \vspace{3pt}
  Non-Retrievable; \\ Entry points from accurate search.
  \vspace{3pt} \end{minipage}
& \begin{minipage}{4.8cm} \vspace{3pt}
  \textit{"AI significance"}
  \vspace{3pt} \end{minipage} \\ \hline

\(R\) 
& Relationship 
& \begin{minipage}{4.5cm} \vspace{3pt}
  Connections between entities represented as nodes. Acts as connector nodes and secondary retrievable node.
  \vspace{3pt} \end{minipage}
& \begin{minipage}{3.5cm} \vspace{3pt}
  Retrievable; \\ Non-Entry points
  \vspace{3pt} \end{minipage}
& \begin{minipage}{4.8cm} \vspace{3pt}
  \textit{"Hinton received the Nobel Prize."}
  \vspace{3pt} \end{minipage} \\ \hline

\(N\) 
& Entity 
& \begin{minipage}{4.5cm} \vspace{3pt}
  Named entities such as people, places, or concepts.
  \vspace{3pt} \end{minipage}
& \begin{minipage}{3.5cm} \vspace{3pt}
  Non-Retrievable; \\ Entry points from accurate search..
  \vspace{3pt} \end{minipage}
& \begin{minipage}{4.8cm} \vspace{3pt}
  \textit{"Hinton," "Nobel Prize"}
  \vspace{3pt} \end{minipage} \\ \hline

\end{tabular}
}
\vspace{10pt}
\caption{Node Types in the heterograph}
\label{tab:node_types_minipage}
\end{table}

\subsection{K-core \& Betweenness centrality}

In this subsection, we present the methodology for identifying important entities and generating their attribute summaries, ensuring alignment with the mathematical framework established in the main text.

The selection of important entities, denoted as \(N^*\), is based on two fundamental structural graph metrics: \(K\)-core decomposition and betweenness centrality. These metrics collectively ensure that the selected nodes are not only structurally integral but also play a pivotal role in facilitating information flow.

The \(K\)-core decomposition, denoted as \(K(\mathcal{G}^1)\), identifies nodes within densely connected subgraphs, ensuring that selected entities contribute significantly to the structural cohesion of the graph. Meanwhile, betweenness centrality, denoted as \(B(\mathcal{G}^1)\), highlights nodes that serve as critical intermediaries between different regions of the graph, capturing entities essential for information dissemination.

The process of identifying important entities follows the steps outlined in Algorithm~\ref{alg:entity_selection}.

\begin{algorithm}[h]
    \caption{Identification of Important Entities}
    \label{alg:entity_selection}
    \begin{algorithmic}
        \STATE \textbf{Input:} Graph $\mathcal{G}^1 = (\mathcal{V}, \mathcal{E})$
        \STATE \textbf{Output:} Important entity set $N^*$
        
        \STATE \textbf{Step 1: Compute \(K\)-core decomposition}
        \STATE Compute the core threshold:
        \[
        k_{\text{default}} = \lfloor \log(|\mathcal{V}|) \times \left( \frac{\sum_{v \in \mathcal{V}} \deg(v)}{|\mathcal{V}|} \right)^{1/2} \rfloor
        \]
        \STATE Extract the \(K\)-core subgraph:
        \[
        K(\mathcal{G}^1) = \{ v \in \mathcal{V} \mid \deg_{\mathcal{G}^1}(v) \geq k_{\text{default}} \}
        \]

        \STATE \textbf{Step 2: Compute betweenness centrality}
        \FOR{each $v \in \mathcal{V}$}
            \STATE Approximate betweenness centrality using shortest-path sampling:
            \[
            b(v) = \text{betweenness\_centrality}(\mathcal{G}^1, k=10)
            \]
        \ENDFOR

        \STATE Compute the average betweenness centrality:
        \[
        \bar{b} = \frac{\sum_{v \in \mathcal{V}} b(v)}{|\mathcal{V}|}
        \]
        \STATE Compute the scale factor:
        \[
        \text{scale} = \lfloor \log_{10}(|\mathcal{V}|) \rfloor
        \]

        \STATE \textbf{Step 3: Select important nodes}
        \FOR{each $v \in \mathcal{V}$}
            \IF{$b(v) > \bar{b} \times \text{scale}$}
                \STATE Add $v$ to $B(\mathcal{G}^1)$
            \ENDIF
        \ENDFOR
        
        \STATE Compute the final set of important entities:
        \[
        N^* = K(\mathcal{G}^1) \cup B(\mathcal{G}^1)
        \]
        
        \STATE \textbf{Return} $N^*$
    \end{algorithmic}
\end{algorithm}

\subsection{Semantic Matching within Community}

To establish meaningful semantic relationships among high-level element nodes, we propose the \textit{Semantic Matching within Community} algorithm. This algorithm ensures that entities with strong semantic similarities are connected within their respective communities. The motivation behind this approach is to organically integrate \( H \) nodes into the graph structure by establishing connections with semantically related nodes within the same community. Formally, the process is summarized in Algorithm~\ref{alg:semantic_matching}.

The algorithm begins by identifying nodes that belong to three specific categories: structure nodes (\( S \)), attribute nodes (\( A \)), and high-level nodes (\( H \)). These nodes are collectively defined as:

\[
\mathcal{V}_{\{S, A, H\}} = \{ v \in \mathcal{V} \mid \psi(v) \in \{S, A, H\} \}
\]

Since these nodes exhibit inherent semantic relationships, we cluster them based on their embeddings, which capture their contextual meaning. To partition the nodes into semantically similar groups, we apply the K-means clustering algorithm \cite{macqueen1967some} to the embedding representations of \( \mathcal{V}_{\{S, A, H\}} \).

which balances computational efficiency and granularity. This clustering process results in a partitioning of nodes into \( K \) semantic clusters, denoted as \( \mathcal{S}_k \), where each cluster contains nodes with closely related semantic representations.

After clustering, the algorithm establishes edges between semantically related nodes within the same community. Communities are predefined structural subgroups in the graph, denoted as \( \mathcal{C}_n \), ensuring that local relationships are preserved. For each community-cluster pair, semantic edges are introduced between nodes in \( \mathcal{V}_{\{S, A\}} \) and nodes in \( \mathcal{V}_H \). Specifically, for any node pair \( (v, v') \), where \( v \in \mathcal{V}_{\{S, A\}} \) and \( v' \in \mathcal{V}_H \), an edge \( e_h(v, v') \) is established if both nodes belong to the same community and the same semantic cluster.

\begin{algorithm}[tb]
    \caption{Semantic Matching within Community}
    \label{alg:semantic_matching}
    \begin{algorithmic}
        \STATE \textbf{Input:} Graph $\mathcal{G} = (\mathcal{V}, \mathcal{E})$, node embeddings $\Phi(\mathcal{V})$, community partition $\{\mathcal{C}_n\}$
        \STATE \textbf{Output:} Semantic edges $\mathcal{E}_h$

        \STATE \textbf{Step 1: Select high-level element nodes}
        \STATE Extract nodes with labels $S$, $A$, or $H$:
        \[
        V_{\{S, A, H\}} = \{ v \in \mathcal{V} \mid \psi(v) \in \{S, A, H\} \}
        \]

        \STATE \textbf{Step 2: Apply K-means clustering to node embeddings}
        \STATE Set number of clusters:
        \[
        K = \sqrt{|\mathcal{V}_{\{S, A, H\}}|}
        \]
        \STATE Perform K-means clustering on $\mathcal{V}_{\{S, A, H\}})$, obtaining clusters $\{\mathcal{S}_k\}$

        \STATE \textbf{Step 3: Establish semantic edges within communities}
        \FOR{each community $\mathcal{C}_n$}
            \FOR{each cluster $\mathcal{S}_k$}
                \STATE Identify nodes within the community and cluster:
                \[
                \mathcal{V}_{\mathcal{C}_n, \mathcal{S}_k} = \mathcal{V}_{\{S, A, H\}} \cap C_n \cap S_k
                \]
                \FOR{each pair $(v, v')$ where $v \in \{S, A\}, v' \in H$}
                    \STATE Add semantic edge:
                    \[
                    e_h(v, v') \in \mathcal{E}_h
                    \]
                \ENDFOR
            \ENDFOR
        \ENDFOR

        \STATE \textbf{Return} $\mathcal{E}_h$
    \end{algorithmic}
\end{algorithm}

By integrating semantic matching within community constraints, this algorithm enhances the structural integrity of the graph while maintaining computational feasibility. The choice of K-means clustering efficiently groups nodes with similar semantic properties, while the enforcement of community constraints ensures that edges are only formed between nodes that naturally belong to the same substructure. Consequently, the proposed method balances semantic consistency and graph locality, making it well-suited for applications requiring structured knowledge representation and retrieval.

\subsection{Dual Search}

To efficiently locate relevant entry points within the Hetero Graph \( \mathcal{G} \), we propose the \textit{Dual Search} algorithm, which integrates exact matching on structured nodes and vector similarity search on rich information nodes. This hybrid approach ensures a balance between precision and recall by leveraging both symbolic and dense representations. The core idea is to utilize exact string matching for well-structured nodes while employing approximate nearest neighbor search for nodes containing rich contextual information. By doing so, the algorithm improves both retrieval accuracy and robustness to query variations.

Given a query, a LLM extracts a set of relevant entities, denoted as \( N^q \), and embeds the query into a vector representation \( \mathbf{q} \). Entry points in the graph are then determined by:

\[
\mathcal{V}_{\text{entry}} = \{v \in \mathcal{V} \mid \Phi(v, N^q, \mathbf{q})\},
\]

where the condition function \( \Phi(v, N^q, \mathbf{q}) \) determines whether a node qualifies as an entry point:

\[
\Phi(v, N^q, \mathbf{q}) = 
\begin{cases} 
v \in \mathcal{V}_{\{N, O\}} \land \mathcal{M}(N^q, v), \\ 
v \in \mathcal{V}_{\{S, A, H\}} \land \mathcal{R}(\mathbf{q}, v, k).
\end{cases}
\]

Here, the exact matching function \( \mathcal{M}(N^q, v) \) returns true if node \( v \) matches one of the extracted entities in \( N^q \). This ensures that titles or named nodes such \(\mathcal{V}_{ N, O }\) are retrieved deterministically. Meanwhile, the similarity-ranking function \( \mathcal{R}(\mathbf{q}, v, k) \) applies HNSW, selecting the top-\(k\) most similar nodes to \( \mathbf{q} \) in vector space. This is particularly beneficial for context-rich nodes for \( v \in \mathcal{V}_{S, A, H} \), which may not contain exact query terms but remain semantically relevant.

One of the major strengths of Dual Search is its ability to mitigate errors from noisy or ambiguous queries. Since \( N \) and \( O \) nodes serve exclusively as entry points and are not directly retrieved, any errors in exact matching do not propagate into the retrieval results. Furthermore, Personalized PageRank (PPR) is applied to refine retrieval by selecting highly weighted cross nodes—nodes that are both structurally significant and semantically aligned with multiple entry points. This graph-based expansion helps compensate for potential errors in entity extraction and ensures robustness against query variations. Another advantage of this approach is its scalability. By employing HNSW, the algorithm avoids the computational overhead of exhaustive similarity searches, making it efficient even on large-scale graphs. 

\newpage
\section{Case Study}
\label{case study}
\begin{table}[H]
\fontsize{11pt}{11.5pt}
\selectfont
\centering

\begin{tabular}{|m{2cm}|m{14cm}|}

\hline
\textbf{Question1}\vspace{0.1em}
& 
\vspace{0.2em}
Why isn’t there public transport on Christmas day in the UK? 
\vspace{0.2em}
\\ 
\hline

\textbf{NodeRAG} & 
\vspace{0.5em}
The absence of public transport on Christmas Day in the UK can be attributed to a combination of historical, cultural, and economic factors. \textcolor[RGB]{0,139,0}{\textbf{Historically}, public transport services were more prevalent on Christmas Day \textbf{before World War II},} as many people worked on that day and required transportation. However, over the years, societal norms have shifted, leading to a significant reduction in demand for public transport services on Christmas Day. 

\vspace{0.3em}

\textcolor[RGB]{0,139,0}{\textbf{In contemporary society}, Christmas is widely recognized as a time for family gatherings and celebrations, which diminishes the need for public transport.} Many people choose to spend the day at home or with relatives, resulting in a lower demand for transit services. Consequently, transport companies have adapted to this change by suspending services on Christmas Day, allowing transit workers to enjoy the holiday with their families as well.

\vspace{0.3em}

Additionally, \textcolor[RGB]{0,139,0}{the \textbf{operational costs} associated with running public transport services on Christmas Day can be significant}. With minimal passenger demand, the financial viability of providing such services becomes questionable. Railways and other transport providers often use this time for maintenance and engineering work, taking advantage of the reduced disruption to carry out necessary improvements.
\vspace{0.3em}
 \\ \hline
\vspace{0.3em}
\textbf{GraphRAG} & 
\vspace{0.5em}
Public transport in the UK is significantly limited on Christmas Day due to a combination of historical, cultural, and operational factors. Traditionally, Christmas Day is a public holiday in the UK, and many businesses, including public transport services, close to allow employees to spend time with their families and celebrate the holiday. This practice is rooted in the cultural significance of Christmas, which emphasizes family gatherings and community celebrations.

\vspace{0.3em}

The limited operation of public transport on Christmas Day is also influenced by the low demand for services. Historically, fewer people travel on this day compared to other holidays or weekends, as many individuals prefer to stay home or visit family. As a result, transport providers often find it economically unfeasible to run full services when the number of passengers is expected to be minimal.

\vspace{0.3em}

In recent years, some transport services have begun to offer limited services on Christmas Day, particularly in major cities like London, where there may be a demand for travel to events or gatherings. However, these services are typically not as extensive as those available on other public holidays or regular days. Overall, the combination of cultural practices, low demand, and operational considerations contributes to the lack of public transport on Christmas Day in the UK.
\vspace{0.3em}
 \\ \hline

\textbf{LightRAG} & 
\vspace{0.5em}
In most parts of the world, official or religious holidays are taken very seriously. On such days, employers typically have to pay higher rates to staff who are willing to work, but that effort often doesn't pay off due to low customer turnout. Families tend to arrange their travel plans by Christmas Eve to avoid the need for transportation on Christmas Day. Christmas avoids unnecessary travel because it's typically a day filled with home activities, and it allows workers in various roles to have a day off. The lack of public demand and the staff's resistance contributed to the lack of services on Christmas Day. Christmas Day is also an opportune time to perform engineering work on railways with minimal disruption. The railroad industry is capable of completing extensive projects during this period, thanks to thorough planning and pre-staging. 
\vspace{0.3em}
\\ \hline

\textbf{LLM \newline Decision} & 
\vspace{0.5em}
ORDER: \{``1": ``GraphRAG", ``2": ``NodeRAG"\}

Both answers provide a comprehensive explanation, but \textcolor[RGB]{205,173,0}{answer 2 adds \textbf{historical context} about public transport before World War II and mentions the use of the holiday for maintenance work, offering slightly \textbf{more depth}}.

\vspace{0.3em}

ORDER: \{``1": ``LightRAG", ``2": ``NodeRAG"\}

Both answers provide a comprehensive explanation, but \textcolor[RGB]{205,173,0}{answer 2 adds \textbf{historical context} about the change in public transport services since World War II, which \textbf{enriches the explanation}. Therefore, answer 2 is slightly better}.
\vspace{0.3em}
 \\ \hline
\end{tabular}
\vspace{5pt}
\caption{Sample Table with Multi-line Text}
\end{table}

\begin{table}[H]
\centering
\begin{tabular}{|m{2cm}|m{14cm}|}
\hline
\textbf{Question2} \vspace{0.1em}& 
\vspace{0.2em}
What are rabbit hides used for in Minecraft?
\vspace{0.2em}\\ 
\hline

\textbf{NodeRAG} & 
\vspace{0.5em}
In Minecraft, rabbit hides serve a couple of useful purposes, particularly in crafting and resource management. \textcolor[RGB]{0,139,0}{One of the primary uses of rabbit hides is to craft leather}. Players can combine four rabbit hides in a 2x2 crafting grid to create one piece of leather. This is especially beneficial in Survival Multiplayer (SMP) scenarios where finding other animals for leather can be challenging.

\vspace{0.5em}

Additionally,  \textcolor[RGB]{0,139,0}{rabbits can drop a unique item known as the \textbf{``rabbit's foot"}, which is an important ingredient for brewing potions. The rabbit's foot is specifically used in the crafting of the \textbf{Potion of Leaping}}, which grants players the ability to jump higher for a limited time. Thus, rabbit hides and their associated drops contribute significantly to crafting and potion-making in the game, enhancing the overall gameplay experience.
\vspace{0.5em}
 \\ \hline

\textbf{GraphRAG} & 
\vspace{0.5em}
In Minecraft, rabbit hides are primarily used to craft leather. Players can combine four rabbit hides in a 2x2 crafting grid to create one piece of leather. This crafting option is particularly useful for players who may not have access to cows, which are the traditional source of leather in the game. 

\vspace{0.5em}

Additionally, rabbit hides can be beneficial in multiplayer servers that utilize entity clears, as finding animals can be challenging. By using rabbit hides to craft leather, players have an alternative method to obtain this valuable resource, which is essential for crafting various items, including armor and books.
\vspace{0.5em} 
 \\ \hline

\textbf{LightRAG} & 
\vspace{0.5em}
In Minecraft, Rabbit Hides are primarily used as a crafting resource. They can be combined with other materials to create leather, which is an essential item in the game. Leather is used for crafting various items, including armor (like leather armor), books, item frames, and more. 

\vspace{0.5em}

The process of obtaining Rabbit Hides involves hunting rabbits, which can be found in various biomes. Once collected, players can use the hides in their crafting recipes to enhance their gameplay experience through resource management and item creation. This highlights the importance of Rabbit Hides within the broader crafting mechanics of Minecraft.
\vspace{0.5em}
\\ \hline

\textbf{LLM \newline Decision} & 
\vspace{0.5em}
ORDER: \{``1": ``TGRAG", ``2": ``GraphRAG"\}

Both answers are truthful, but \textcolor[RGB]{205,173,0}{answer 1 provides more \textbf{detailed information} about the uses of rabbit hides, including the additional use of \textbf{rabbit's foot for potion-making} Therefore, answer 1 is better.}

\vspace{0.5em}

ORDER: \{``1": ``TGRAG", ``2": ``LightRAG"\}

Both answers are truthful, but \textcolor[RGB]{205,173,0}{answer 1 provides more \textbf{detailed information} about the uses of rabbit hides, including the crafting of leather and \textbf{the rabbit's foot for potions}. Therefore, answer 1 is better.}
\vspace{0.5em}
 \\ \hline

\end{tabular}
\vspace{0.5em}
\caption{Sample Table with Multi-line Text}
\end{table}

\newpage
\section{Overview Prompting in NodeRAG}
\label{Prompting in NodeRAG}

\begin{figure}[H]
    \centering
    \resizebox{1.05\textwidth}{!}{\includegraphics{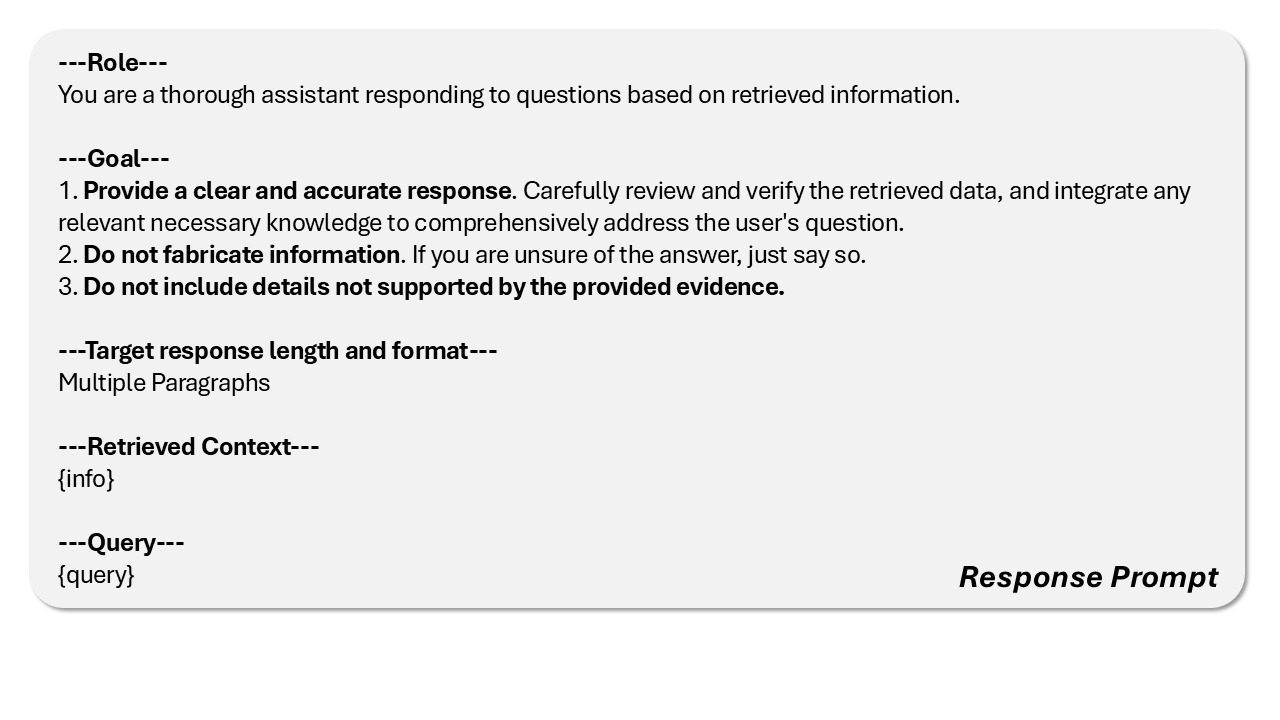}}
    \label{fig:Prompt1}
\end{figure}

\begin{figure}[H]
    \centering
    \resizebox{1.05\textwidth}{!}{\includegraphics{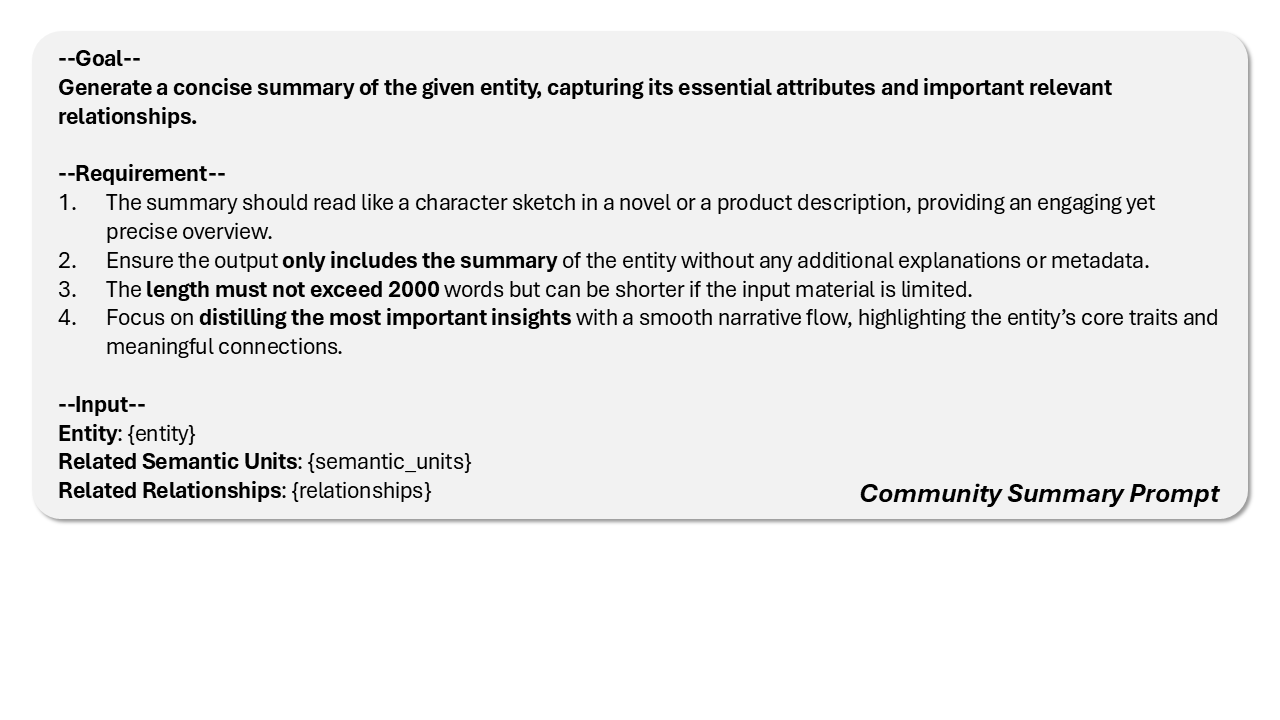}}
    \label{fig:Prompt2}
\end{figure}

\begin{figure}[H]
    \centering
    \resizebox{1.05\textwidth}{!}{\includegraphics{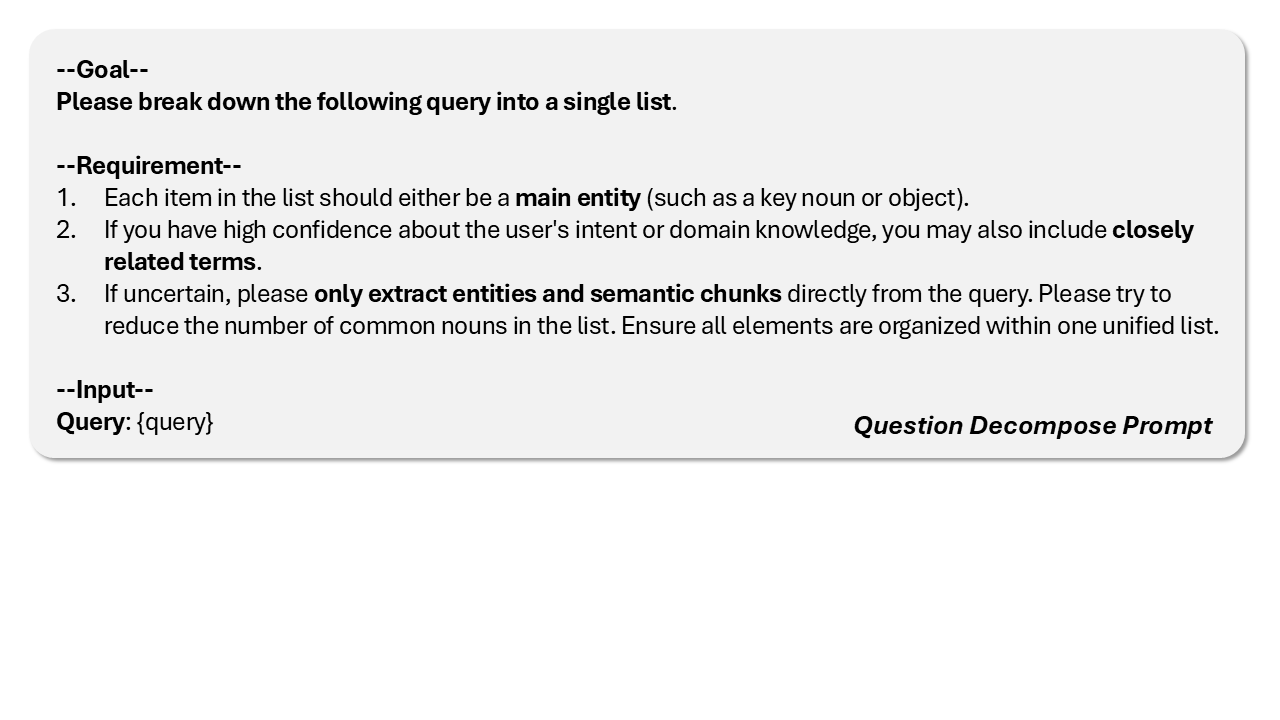}}
    \label{fig:Prompt3}
\end{figure}

\begin{figure}[htbp]
    \centering
    \resizebox{1.05\textwidth}{!}{\includegraphics{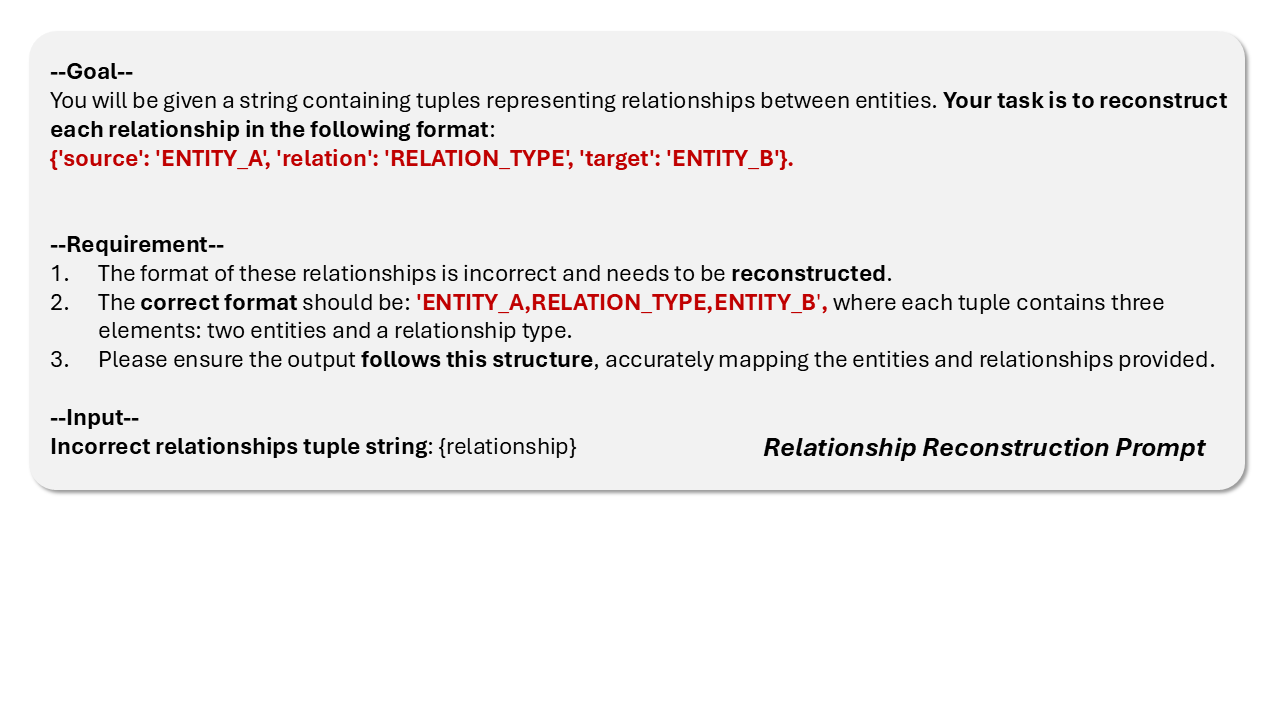}}
    \label{fig:Prompt4}
\end{figure}

\begin{figure}[htbp]
    \centering
    \includegraphics[width=\textwidth, height=0.99\textheight, keepaspectratio]{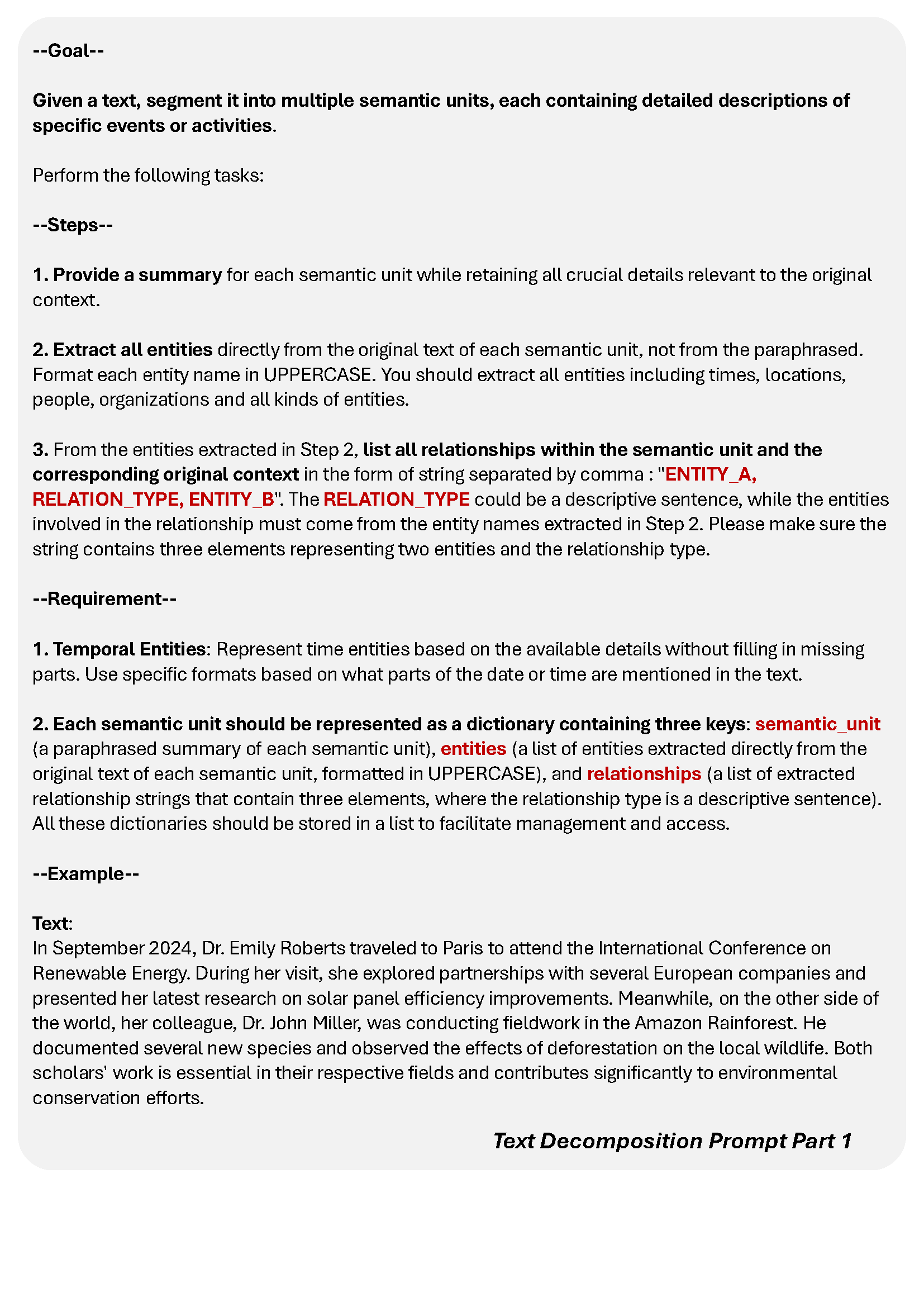}
    \label{fig:Prompt5}
\end{figure}

\begin{figure}[htbp]
    \centering
    \includegraphics[width=\textwidth, height=0.99\textheight, keepaspectratio]{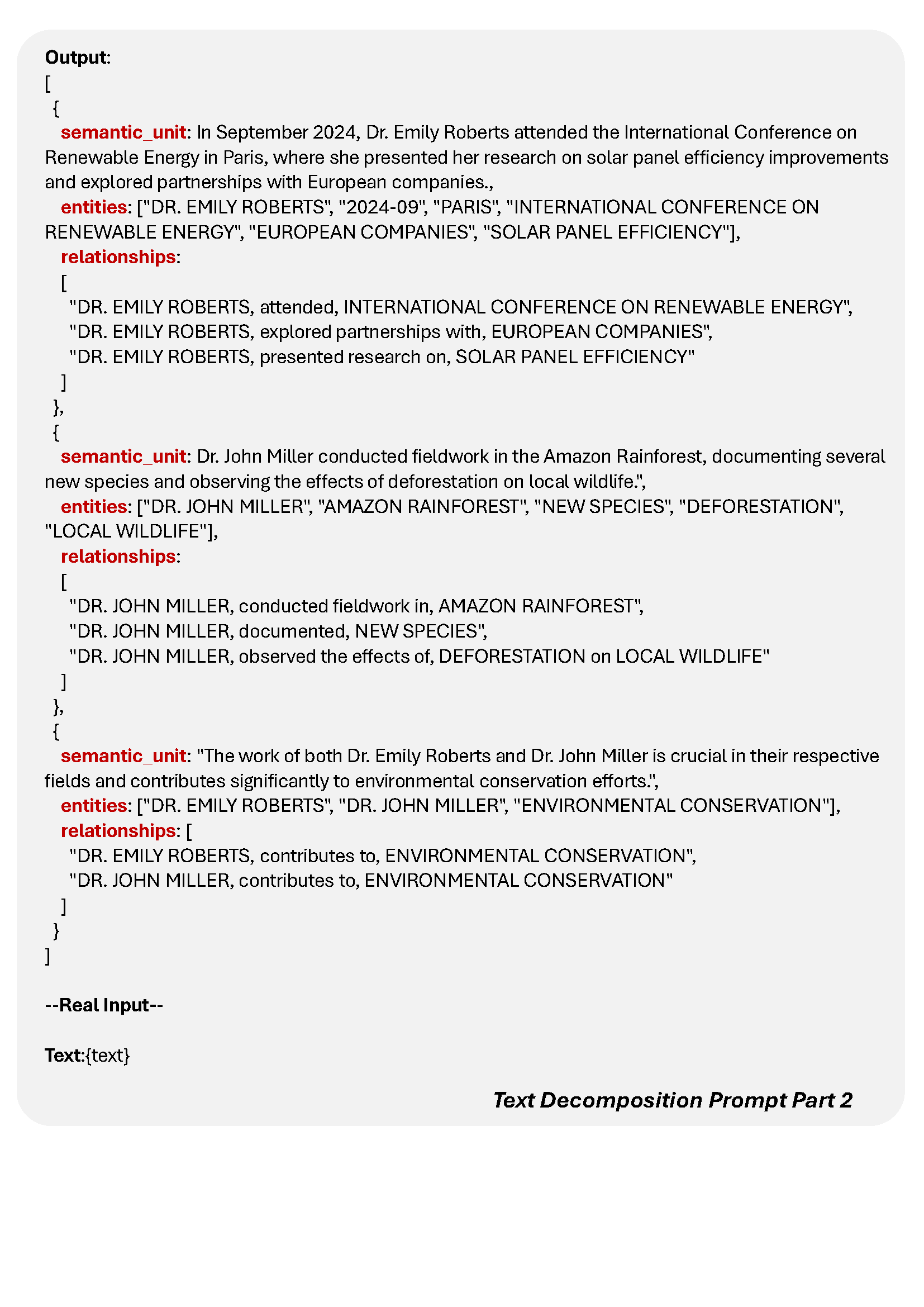}
    \label{fig:Prompt6}
\end{figure}

\end{document}